%% file: arxiv.tex
\documentclass[sigconf,bookmarks=false]{acmart}
\AtBeginDocument{%
  \providecommand\BibTeX{{%
    \normalfont B\kern-0.5em{\scshape i\kern-0.25em b}\kern-0.8em\TeX}}}

\copyrightyear{2022}
\acmYear{2022}
\setcopyright{acmcopyright}
\acmConference[WWW '22] {Proceedings of the ACM Web Conference 2022}{April 25--29, 2022}{Virtual Event, Lyon, France.}
\acmBooktitle{Proceedings of the ACM Web Conference 2022 (WWW '22), April 25--29, 2022, Virtual Event, Lyon, France}
\acmPrice{15.00}
\acmISBN{978-1-4503-9096-5/22/04}
\acmDOI{10.1145/3485447.3512159}

\usepackage{bm}
\usepackage{xspace}
\usepackage{enumitem}
\usepackage[textsize=tiny]{todonotes}
\makeatletter
\DeclareRobustCommand\onedot{\futurelet\@let@token\bmv@onedotaux}
\def\bmv@onedotaux{\ifx\@let@token.\else.\null\fi\xspace}
\def\eg{\emph{e.g}\onedot} 
\def\ie{\emph{i.e}\onedot} 
 
\def\etc{\emph{etc}\onedot} \def\vs{\emph{vs}\onedot}
\def\wrt{w.r.t\onedot}

\begin{document}

\title{Graph-adaptive Rectified Linear Unit for Graph Neural Networks}

\author{Yifei Zhang}
\email{yfzhang@cse.cuhk.edu.hk}
\affiliation{%
\institution{The Chinese University of Hong Kong}
  \city{Hong Kong SAR}
  \country{China}
}

\author{Hao Zhu}
\email{allenhaozhu@anu.edu.au}
\affiliation{%
  \institution{Australian National University and Data61/CSIRO}
  \city{Canberra}
  \country{Australia}
}

\author{Ziqiao Meng}
\email{zqmeng@cse.cuhk.edu.hk}
\affiliation{%
\institution{The Chinese University of Hong Kong}
  \city{Hong Kong SAR}
  \country{China}
}

\author{Piotr Koniusz}
\email{piotr.koniusz@data61.csiro.au}
\affiliation{%
  \institution{Data61/CSIRO and Australian National University}
  \city{Canberra}
  \country{Australia}
}

\author{Irwin King}
\email{king@cse.cuhk.edu.hk}
\affiliation{%
  \institution{The Chinese University of Hong Kong}
  \city{Hong Kong SAR}
  \country{China}
}


\input{Content/abstract}


\maketitle
\input{Content/Introduction}
\input{Content/RelatedWork}
\input{Content/Preliminary}
\input{Content/Method}
\input{Content/Experiment}
\input{Content/Conclusion}
\begin{acks}
We would like to thank the anonymous reviewers for their valuable comments. The work described in this paper was partially supported by the National Key Research and Development Program of China (No. 2018AAA0100204) and CUHK 2300174, Collaborative Research Fund (CRF), No. C5026-18GF. We also acknowledge the in-kind support from the Amazon AWS Machine Learning Research Award (MLRA) 2020.
\end{acks}
\bibliographystyle{ACM-Reference-Format}
\balance
\bibliography{arxiv}
\end{document}

%% file: Content/abstract.tex
\begin{abstract}
Graph Neural Networks (GNNs) have achieved remarkable success by extending traditional convolution to learning on non-Euclidean data. 
The key to the GNNs is adopting the neural message-passing paradigm with two stages: aggregation and update. The current design of GNNs considers the topology information in the aggregation stage. However, in the updating stage, all nodes share the same updating function. The identical updating function treats each node embedding as  i.i.d. 
random variables and thus ignores the implicit relationships between neighborhoods, which limits the capacity of the GNNs. The updating function is usually implemented with a linear transformation followed by a non-linear activation function. To make the updating function topology-aware, we inject the topological information into the non-linear activation function and propose Graph-adaptive Rectified Linear Unit (GReLU), which is a new parametric activation function incorporating the neighborhood information in a novel and efficient way. The parameters of GReLU are obtained from a hyperfunction based on both node features and the corresponding adjacent matrix. To reduce the risk of overfitting and the computational cost, we decompose the hyperfunction as two independent components for nodes and features respectively. We conduct comprehensive experiments to show that our plug-and-play GReLU method  is efficient and effective given different GNN backbones and various downstream tasks.
\end{abstract}

%% file: Content/Introduction.tex
\begin{figure}[t]
\centering
\includegraphics[width=\columnwidth]{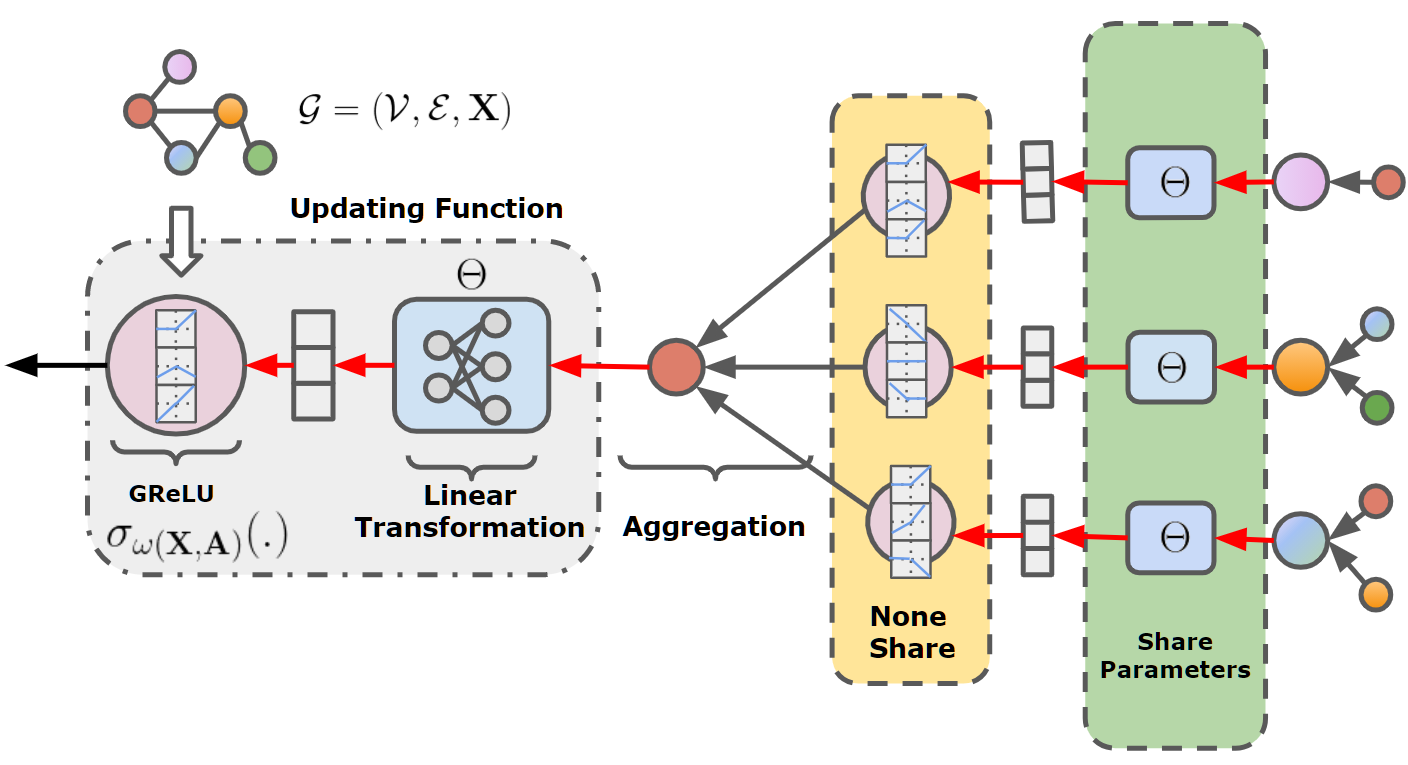}
\caption{Plugging GReLU in message passing with three channels.  
The grey box with dash lines shows details of the updating function which consists of a linear transformation (the blue box) and a non-linear activation function (the pink circle). The black and red arrows indicate the aggregation and updating processes. Note that the linear transformation is shared across different nodes while the activation functions are both node- and channel-specific which are adaptive to both features and the adjacency matrix.}
\label{fig:main}
\vspace{-0.3cm}
\end{figure}
\section{Introduction}
Deep learning has revolutionized many machine learning tasks in recent years, ranging from computer vision to speech and natural language understanding~\cite{lecun2015deep}. The data in these tasks is typically represented in the Euclidean space. However, there is an increasing number of applications where data is generated from non-Euclidean domains and is represented as graphs with complex relationships between objects~\cite{lee2020news,lim2020stp,wang2020efficient,yang2021discrete,yang2020featurenorm,fu2020magnn,IDX}. By defining the convolution operators on the graph, graph neural networks (GNNs) extend convolution neural networks (CNNs) from the image domain into the graph domain. Thus, we can still adapt may tools from CNNs to GNNs, \ie, ReLU~\cite{nair2010rectified}, Dropout~\cite{srivastava2014dropout}, or residual link~\cite{he2016deep}. 

GNN is now capable of mining the characteristics of the graph by adopting the so-called message passing (MP) framework, which iteratively aggregates and updates the node information following the edge path. Embracing the MP as a modeling tool, recent years witnessed an unprecedented achievement in downstream tasks, such as node/graph classification~\cite{DBLP:conf/cikm/SongMZK21,DBLP:journals/corr/zixing} and link prediction~\cite{chen2021modeling,chen2021attentive}. However, proven by many recent works, an inappropriately designed propagation step in MP violates the homophily assumption held by many real-world graphs, consequently causing critical performance degradation~\cite{DBLP:conf/iclr/supergat,DBLP:conf/nips/BeyondHomophily}.

The aforementioned nuisances motivate us to refine the MP and enhance the generalization ability of GNN. One promising line of exploration is to dynamically consider the contribution of each node during MP. Attention mechanisms are a good pointer as they have been explored in~\cite{velivckovic2017graph,DBLP:conf/iclr/supergat}. One of the benefits of attention mechanisms is that they let network focus on the most relevant information to propagate by assigning the weight to different nodes accordingly \eg, works~\citep{velivckovic2017graph,DBLP:conf/iclr/supergat} show that propagating information by adaptive transformation yields impressive performance across several inductive benchmarks.

However, the current design of adaptive transformation is inefficient. Our reasoning arises from two critical standpoints. Firstly, no topology information has been considered in inference of the adaptive weight for each input node, thus, failing to capture the homophily information of the graph~\cite{DBLP:conf/iclr/supergat}. Secondly, we suspect the limitation in improving model generalization ability and accuracy is due to sharing the adaptive weight across channels for each input node. It may be undesirable to transform all channels with no discrimination as some channels are less important than others.

Motivated by the issues above, we develop GReLU, a topology-aware adaptive activation function, amenable to any GNN backbone. GReLU refines the information propagation and improves the performance on node classification tasks. As illustrated in Figure~\ref{fig:main}, our mechanism is comprised of two components improving the training procedure of GNNs. On the one hand, the GReLU inherits the topology information by taking features and graphs as input. On the other hand, unlike standard rectifiers being the same function \wrt all inputs, GReLU assigns a set of parameters to dynamically control the piecewise non-linear function of each node and channel. 
% In principle, making the non-linear piecewise function adaptive to the input. Meanwhile, GReLU is computationally efficient since it has a negligible computational cost. Overall, the main contributions of this work include:
Such a topology-adaptive property provides a better expressive power than the topology-unaware counterpart. Meanwhile, GReLU is computationally efficient \eg,  a two-layer GCN with GReLU and a two-layer GAT~\cite{velivckovic2017graph} have similar computational complexity, while the former is able to outperform GAT in node classification. 

\vspace{0.1cm}
The main contributions of this work are listed below:

\renewcommand{\labelenumi}{\roman{enumi}.}
\vspace{-0.4cm}
\hspace{-1cm}
\begin{enumerate}[leftmargin=0.6cm]
%\item We formulate the FSL problem as to generate dynamic classifiers.
\item  We incorporate the topology information in the updating function of GNNs by introducing a topology-aware activation function in the updating stage of message passing, which, to our knowledge, is novel.
    \item We propose a well-designed graph adaptive rectified linear unit for graph data, which can be used as a plug-and-play component in various types of spatial and spectral GNNs.
    \item We evaluate the proposed methods on various downstream tasks. We empirically show that the performance of various types of GNNs can be generally improved with the use of our GReLU.
\end{enumerate}

%% file: Content/RelatedWork.tex
\section{Related Works}
\subsection{Graph Neural Networks}
 GNNs can be categorized into spectral  or spatial domain type~\cite{9046288}. The spectral domain is exploring GNNs design from the perspective of graph signal processing \ie, the Fourier transform and the eigen-decomposition of the Laplacian matrix~\cite{defferrard2016convolutional} are used to define the convolution step. One can extract any frequency to design a graph filter with the theoretical guarantees. However, these models suffer from three main drawbacks: 1) a large amount of computational burden is induced by the eigen-decomposition  of the given graph; 2) spatial non-localization; 3) the filter designed for a given graph cannot be applied to other graphs. Spatial GNNs generalize the architecture of GNNs by introducing the message passing mechanism, which is aggregating neighbor information based on some spatial characteristics of the graph such as adjacency matrix, node features, edge features, \etc~\cite{hamilton2017inductive}. Due to its high efficiency and scalability, spatial GNNs have become much more popular in recent works. The most representative model is graph attention network (GAT)~\cite{velivckovic2017graph}, which weights the node features of different neighbors by using the attention mechanism in the aggregation step. However, for each node feature, GAT assigns the node a single weight  shared across all channels, which ignores the fact that different channels may exhibit different importance.

\subsection{Activation Functions}
Activation functions have been long studied over the entire history of neural networks. One of the few important milestones is introducing the Rectified Linear Unit (ReLU)~\cite{nair2010rectified} into the neural network. Due to its simplicity and non-linearity, ReLU became the standard setting for many successful deep architectures. A series of research has been proposed to improve ReLU. 
Two variants of ReLU are proposed which adopt non-zero slopes $\alpha$ for the input less than zero: Absolute value rectification choose $\alpha = -1$~\cite{jarrett2009best}; LeakyReLU fixes $\alpha$ to a small value~\cite{leakyrelu}. PReLU~\cite{prelu} and \cite{chen2020dynamic} takes a further step by making the nonzero slope a trainable parameter. Maxout generalizes ReLU further, by dividing the input into groups and outputting the maximum~\cite{goodfellow2013maxout}. Since ReLU is non-smooth, some smooth activation functions have been developed to address this issue, such as soft plus~\cite{softplus}, ELU~\cite{elus}, SELU~\cite{klambauer2017self},  Misc~\cite{misra2019mish}, Exp \cite{ckn} and Sigmoid \cite{koniusz2018deeper}.
These rectifiers are all general functions that can be used in different types of neural networks. None of them has a special design for GNNs by considering the topology information.
% need to point out why we are different.

\subsection{Linear Graph Neural Network}%%confused......
In \cite{wu2019simplifying,pmlr-v115-sun20a,zhu2020simple,zhu2021contrastive,zhu2021refine}, the authors proposed simplified GNNs (SGC and S\textsuperscript{2}GC) by removing the activation function and collapsing the weight matrices between consecutive layers. These works hypothesize that the non-linearity between GCN layers is not critical but that the majority of the benefit arises from the topological smoothing. 
However, removing non-linearity between layers sacrifices the depth efficiency of GNNs. 
Moreover, the ReLU removed from SGC is not topology-aware. Hence, the rectifiers play a less important role in GNN. 
As stated in SGC, the topology information is crucial in the aggregation stage to make GNN work, which also motivates us to incorporate the topology information in the updating stage.

%% file: Content/Preliminary.tex
\section{Preliminaries}
% In this section, we present the notions and preliminary knowledge used in this paper. Without special statement we use the bold upper and bold lower letters to indicate matrix and vector respectively and use unbold lower letters to indicate scalar.
\subsection{Notations} 
\label{notions}
In this paper, a graph with node features is denoted as $\mathcal{G}=(\bm{X}, \bm{A})$, where $\mathcal{V}$ is the vertex set, $\mathcal{E}$ is the edge set, and $\bm{X} \in \mathbb{R}^{N \times C}$ is the feature matrix where $N=|\mathcal{V}|$ is the number of nodes and $C$ is the number of channels. $\bm{A} \in\{0,1\}^{N \times N}$ denotes the adjacency matrix of $\mathcal{G}$ , \ie, the $(i, j)$-th entry in $\bm{A}$, $a_{ij}$, is 1 if there is an edge between $v_{i}$ and $v_{j}$. We denote $\mathcal{N}(i) = \{j | {a}_{ij} = 1\}$ as the set of indexes of neighborhoods of  node $i$. The degree of $v_{i}$, denoted as $d_{i}$, is the number of edges incident with $v_{i}$. For a $d$-dimensional vector $\bm{x} \in \mathbb{R}^{d}$, we use ${x}_{i}$ to denote the $i$-th entry of $\bm{x}$. We use $\bm{x}_{i}$ to denote the row vector of $\bm{X}$ and ${x}_{ij}$ for the $(i, j)$-th entry of $\bm{X}$. 

\subsection{Message Passing (MP)}
\label{msg passing}
The success of spatial GNNs results from applying message passing (also called neighbor aggregation) between nodes in the graph. In the forward pass of GNNs, the hidden state of node $i$ is iteratively updated via aggregating the hidden state of node $j$ from $i$'s neighbors through edge $e_{ij}$. Suppose there are $l$ iterations, the message passing paradigm can be formalized as:
\begin{align}
    &\bm{m}_i^{(l)} \;\;= \text{{AGGREGATE}}^{(l)}\big(\{ \bm{h}^{(l)}_i, \bm{h}^{(l)}_j, \bm{e}_{ij} | j \in \mathcal{N}(i)\}\big),\\
    &\bm{h}_i^{(l+1)} = \sigma\big({\bm{\Theta}^{(l)}}[\bm{h}^{(l)}_i, \bm{m}^{(l)}_i]\big),
    \label{equ:2}
\end{align}
where $\bm{h}_i$ and $\bm{h}_j$ are  hidden states of nodes $i$ and $j$, $\bm{m_i}$ denotes the message that the node $i$ receives from its neighbors $j \in \mathcal{N}(i)$. The updating function in \eqref{equ:2} is modeled as a linear transformation followed by a non-linear activation function $\sigma(\cdot)$ where $\bm{\Theta}\in\mathbb{R}^{C\times(2C)}$ matrix contains learned parameters for the linear transformation and $[\cdot]$ concatenates vectors.
\subsection{Parametric ReLU}
\label{param relu}
The vanilla ReLU is a piecewise linear function, $\bm{y} = \max(\bm{x},0)$, where $\max(\cdot)$ is an element-wise function applied between each channel of the input
$\bm{x}$ and 0. A more generalized way to extend ReLU is to let each channel enjoy different element-wise function:
% \vspace{-0.1cm}
\begin{equation}
    y_c = \sigma_c(x_c)  = \max_{1< k \leq K}\{\alpha^k x_c + \beta^k\},
\label{eq:PRELU}
\end{equation}
where $x_c$ and $y_c$ are the $c$-th channels of $\bm{x}$ and $\bm{y}$ respectively, $k$ is the $k$-th segmentation in \eqref{eq:PRELU}, and $\{(\alpha^k, \beta^k)\}_{k=1}^{K}$ is the set of parameters for the parametric ReLU. Note that when $\alpha^1=0, \beta^1=0, \alpha^2=1, \beta^2=0$, then \eqref{eq:PRELU} reduces to ReLU, whereas maxout~\cite{goodfellow2013maxout} is a special case if $K=2$. Furthermore, instead of learning parameters $(\alpha^k, \beta^k)$ directly, approach \cite{chen2020dynamic} proposes the hyperfunction $\omega(x)$ to estimate sample-specific $(\alpha^k, \beta^k)$ for different channels.
%%Highlight Parametric ReLU is independent.

%% file: Content/Method.tex
\section{Method}
\label{sec:4}
We refer to the proposed adaptive rectified linear unit for GNNs as \textbf{GReLU}. The rest of the sections are organized as follows. Firstly, we introduce the parametric form of GReLU by defining two functions: the hyperfunction that generates parameters and activation function for computing the output. Then, we detail the architecture of hyperfunction of  GReLU, and we link GReLU with GAT from the perspective of  neighborhood  weighting. We finally analyze the time complexity of GReLU and compare it with prior works. 

\subsection{Graph-adaptive Rectified Linear Unit (GReLU)$\!\!\!\!\!\!\!\!$}
\begin{figure}[t]
\centering
\includegraphics[width=0.5\textwidth]{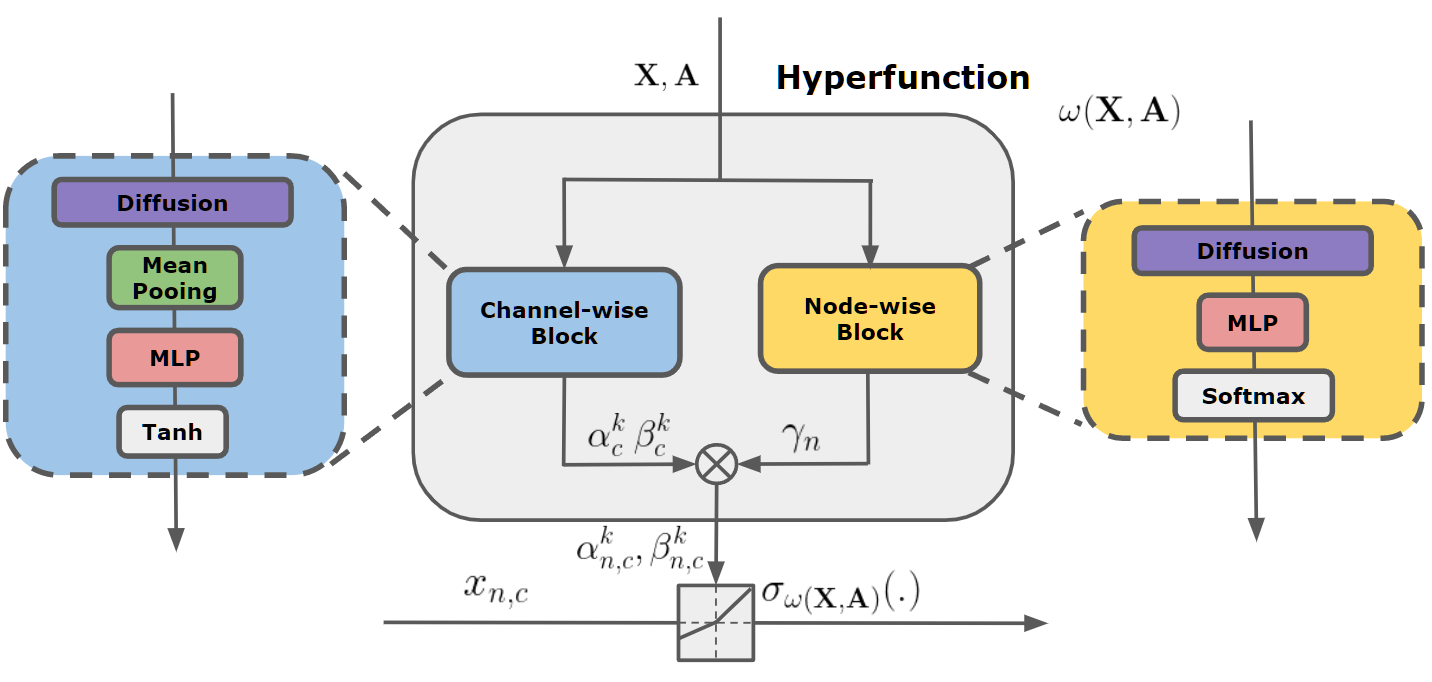}
\caption{The architecture of GReLU. The arrows in the box indicate the flow for computing the parameter for hyperfunction. The arrow outside the box indicates the calculation of GReLU. The channel-wise block produces the $K\times C$ parameters for $C$ channels and $K$ segments. The node-wise block generates $N$ parameters for $N$ nodes. The total $K\times C \times N$ parameters are obtained via computing the outer product of channel-wise and node-wise parameters.}\label{fig:1}
\end{figure}

Given the input feature of node $n$ where $\bm{x}_n\in\mathbb{R}^C$,
%$ = \{x_{nc}\}_{c=1}^{C}$, 
the GReLU is defined as the element-wise function:
% \vspace{-0.1cm}
\begin{equation}
    \sigma_{\omega(\bm{X}, \bm{A})}(x_{nc}) = \max_{1\leq k\leq K}\{\alpha^k_{nc}x_{nc}+\beta^k_{nc}\},
\label{eq:GReLU}
\end{equation}
which consists of two parts:
\begin{itemize}
    \item Hyperfunction $\omega(\bm{X}, \bm{A})$ whose inputs are the adjacency matrix $\bm{A}$ and the node feature matrix $\bm{X}$, and whose output parameters $\{(\alpha^k_{nc}$, $\beta^k_{nc})\}_{k=1}^{K}$ are used by the parametric activation function.
    \item The parametric activation function $\sigma(\cdot)$ generates activation with parameters $\alpha^k_{nc}$.
\end{itemize}
  The hyperfunction encodes the adjacency matrix $\bm{A}$ and node features matrix $\bm{X}$ to determine the parameters of GReLU, adding the topology information to the non-linear activation function. By plugging GReLU into MP, the updating function becomes adaptive, gaining more representational power than its non-adaptive counterpart.

\subsection{Design of Hyperfunction of GReLU}
Note that to fully get GReLU, the hyperfunction needs to produce $K$ sets of parameters $\{(\alpha^k_{nc}, \beta^k_{nc})\}_{k=1}^{K}$ for $C$ channels and $N$ nodes. The total number of parameters needed in GReLU is $2K\times C\times N$. We can simply model the hyperfunction $\omega(\bm{X}, \bm{A})$ as a one-layer graph convolution neural network (GCN) with output dimensions $2K\times C\times N$. However, this implementation results in too many parameters in the hyperfunction, which poses the risk of overfitting, as we observe performance degradation in practice. To solve this issue, we decompose the parameters of nodes from channels by modeling them as two different blocks. One block learns the channel-wise parameters $\alpha^k_{c}, \beta^k_{c}$ and another block learns node-wise parameters $\gamma_n$. The final outputs $(\alpha^k_{nc}, \beta^k_{nc})$ are computed as:
%the product between $(\alpha^k_c, \beta^k_c)$ and $\gamma_n$,
% \vspace{-0.3cm}
\begin{align}
\label{equ:GReLUParameters}
        &\alpha_{nc}^k = \gamma_n\alpha_c^k\;\;\text{and}\;\;\beta_{nc}^k = \gamma_n\beta_c^k.
\end{align}
This factorization limits parameters that need to be learned by the hyperfunction from $2K\times C\times N$ to $2K\times(C+N)$. As a result, the generality of the model and the computational efficiency are improved.

\begin{table}[t]
\caption{Different  Variants of ReLU.}
\centering
  \resizebox{\columnwidth}{!}{
  \begin{tabular}{lllcc}
    \toprule
    \textbf{Type}  & \textbf{Formula} & \textbf{Learnable} & \textbf{Input} & \textbf{Plugin}\\
    \midrule
    ReLU  &  $\max(x, 0)$ & No &$\bm{X}$ & Yes\\
    LReLU & $\max(x, \alpha x)$ & No &$\bm{X}$ & Yes\\
    ELU   &  $\max(x, \alpha (e^x-1))$& No &$\bm{X}$ & Yes\\
    PReLU &  $\max(x, \alpha_p x)$& Yes &$\bm{X}$ & Yes\\
    Maxout &  $\max(w_1x + b_1, w_2x + b_2)$& Yes &$\bm{X}$ & No\\
    \midrule
    GReLU & $ \max_{1\leq k \leq K}\{\alpha^k_{nc}x_{nc} + \beta^k_{nc}\}$ & Yes &$\bm{X},\bm{A}$ & Yes\\
    \bottomrule
    \end{tabular}%
    }
  \label{tab:relus}%
\end{table}
\vspace{0.1cm}\noindent\textbf{Channel-wise Parameters.}
To encode the global context of the graph $\mathcal{G}$ while reducing the number of learnable parameters as much as possible, we adopt the graph diffusion filter $\bm{S}^{PPR}$ to squeeze and extract the topology information from $\bm{A}$ and $\bm{X}$. We propagate input matrix $\bm{X}$ via the fully personalized PageRank scheme:
\begin{equation}
    \bm{E} = \bm{S}^{\mathrm{PPR}}\bm{X}=\alpha\left(\bm{I}_{n}-(1-\alpha) \bm{D}^{-1 / 2} \bm{A} \bm{D}^{-1 / 2}\right)^{-1}\bm{X},
\end{equation}
with teleport (or restart) probability $\alpha \in (0, 1]$, where $\bm{D}$ is the degree matrix and $\bm{D}^{-1 / 2} \bm{A} \bm{D}^{-1 / 2}$ denotes the normalized adjacency matrix $\bm{A}$. We average the diffusion results and transform them into the channel-wise parameters $(\alpha_c^k, \beta_c^k)$ with the use of a linear transformation followed by a tanh function. We normalize $(\alpha_c^k, \beta_c^k)$ to be within $[-1, 1]$ range using $\tanh(\cdot)$ with the purpose of controlling the parameter scale. 
We define  $\bm{h} \in \mathbb{R}^C$ as the average of diffusion results and $\bm{P}\in \mathbb{R}^{2\times K \times C}$ as the channel-wise parameters. The channel-wise block is computed as:
\begin{align}
\label{equ:channelWisedParamters}
        \bm{P} = \tanh(\text{MLP}(\bar{\bm{e}}))\quad \text{where}\quad \bar{\bm{e}} &= \frac{1}{N}\sum_{n=1}^
        N\bm{e}_{n}.
\end{align}
\vspace{0.1cm}\noindent\textbf{Node-wise parameters.}
The node-wise block computes the parameter for each node. We also use the diffusion matrix to capture the graph information. To get the parameter $\gamma_n$ for each node $n$, instead of averaging the diffusion results, we directly squeeze each $\bm{e}$ into one dimension using MLP. To reflect the importance of each node,   softmax is applied to normalize $\gamma_n$ as:
% \vspace{-0.3cm}
\begin{align}
\label{equ:NodeWisedParamters}
        \gamma_{n} = & \frac{\exp(\text{MLP}(\bm{e}_{n}))}{\sum_{n'=1}^N \exp(\text{MLP}(\bm{e}_{n'}))}.
\end{align}

\vspace{0.1cm}\noindent\textbf{Plugging Step into the Updating Function.}
By combining equations \eqref{equ:NodeWisedParamters} and \eqref{equ:channelWisedParamters}, we obtain the full parameters for GReLU via \eqref{equ:GReLUParameters}. We plug GReLU into the $l$-th layer updating function:
\begin{align}
\bm{h}_n^{(l+1)} &= \sigma_n^{(l)}\big({\bm{\Theta}^{(l)}} [\bm{h}_n^{(l)}, \bm{m}_n^{(l)}]\big),
\end{align} 
where $\sigma_n^l(\cdot)$ is GReLU in the $l$-th layer. %The subscript $_{nc}$ indicates the $c$-th channel of $n$-th node. 
As illustrated in Figure~\ref{fig:main}, in the updating function, there is one set of parameters $\bm{\Theta}^{(l)}$ per layer $l$, while GReLU is channel- and nose-adaptive.

\vspace{0.1cm}\noindent\textbf{Application to spectral GNNs.}
We also notice the proposed method is also applicable for Spectral GNNs not only spatial GNNs. Specifically, the graph convolution layer in the spectral domain can be written as a sum of filtered signals followed by an activation function as:
\vspace{-0.2cm}
\begin{equation}
\bm{H}^{(l+1)}=\sigma\Big(\big(\sum_{i=0}^{k} \theta_{i} \lambda_{i} \bm{u}_{i} \bm{u}^{\top}_{i}\big)\bm{H}^{l}\Big)=\sigma(\bm{U} \bm{g}_{\theta}(\bm{\Lambda}) \bm{U}^{\top} \bm{H}^{(l)})
\end{equation}
Here, $\sigma(\cdot)$ is the activation function, $\bm{H}^{(l)}$ is the hidden representation of nodes in the $l$-th layer, $\bm{U}$ contains  eigenvectors of the graph Laplacian $\bm{L} = \bm{D} - \bm{A} $ (or $\bm{L}=\bm{I} - \bm{D}^{-1/2}\bm{A}\bm{D}^{-1/2}$), where $\bm{D} \in \mathbb{R}^{N\times N} $is the diagonal degree matrix with entries, $\lambda_i$ are eigenvalues of $\bm{L}$
and $\bm{g}_{\bm{\theta}}(\cdot)$ is the frequency filter
function controlled by parameters $\bm{\theta}$, where $k$ lowest frequency components are aggregated. Our method can be plugged by simply replacing the $\sigma(\cdot)$ with GReLU.
\begin{figure}
\centering
         \includegraphics[width=0.8\columnwidth]{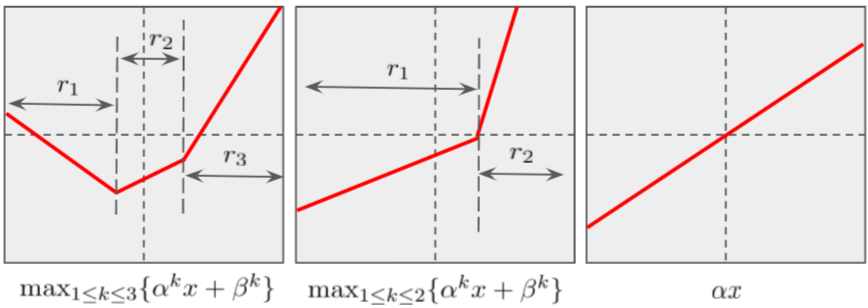}
         \caption{Plots to the left and middle side are the piecewise functions in GReLU with $K=3$ and $K=2$. They cut the input value $x$ in different ranges $r_1$, $r_2$, $r_3$ and re-scale the $x$ \wrt $r$. The plot to the right  is the weighted function used in GAT: the input is re-scaled regardless of the value range of $x$.}
         \label{fig:seg}
\end{figure}
% \begin{table}[t]
% \caption{Summary of Datasets.}
%     \resizebox{0.9\columnwidth}{!}{\begin{tabular}{llllc}
%     \toprule
%     \textbf{DataSet}  & \textbf{Type} & \textbf{\#Nodes} & \textbf{\#Edge} & \textbf{AvgDegree}\\
%     \midrule
%     Cora  &  Citation & 2,708 & 5,429 & 2.00\\
%     CiteSeer & Citation & 3,327 & 4,732 & 1.41\\
%     PubMed   &  Citation & 19,717& 44,338 &2.24\\
%     BlogCatalog &  Social & 5,196 & 171,743& 32.38\\
%     Flickr &  Social &7,575 & 239,738 & 117.05\\
%     \bottomrule
%     \end{tabular}%
%     }
%   \label{tab:datasets}%
% \end{table}
\subsection{Understanding GReLU with Weighted Neighborhood and Attention Mechanism}
\label{sec:udstandgrelu}

Below, we analyze GReLU through the lens of Weighted Neighborhood and Attention Mechanism. To this end, we first introduce the so-call masked attention adopted in the graph attention network (GAT)~\cite{velivckovic2017graph}, which uses the following form:
\begin{equation}
\label{equ:GATForm}
    \begin{aligned}
    \bm{h}'_n = \alpha_n \bm{h}_n\;\; \text{and}\;\;
    \alpha_n = \text{MaskAttention}(\bm{H}, \bm{A}),
\end{aligned}
\end{equation}
where $\alpha_n$ is the attention coefficient assigned to node $n$. Note that \eqref{equ:GATForm} has the linear form of GReLU where the bias parameter and  $\max(\cdot,\cdot)$ is omitted by setting $\beta=0$ and $K=1$. When $K=1$, there is only one segment in the piecewise function of GReLU, the non-linear $\max$ operation  reduces to a linear transformation and thus it can be removed. An illustration is shown in Figure~\ref{fig:seg}. The success of GAT is due to  selecting the important 
node information by weighting them differently during aggregation. While GAT is able to consider the difference of individual nodes, it fails to distinguish the difference between channels. GReLU solves this issue by making the parameter adaptive to both nodes and channels:
\begin{equation}
    \begin{aligned}
    h'_{nc} = \max_{1\leq k \leq K}\{\alpha^k_{nc}h_{nc} + \beta^k_{nc}\}\;\;\text{where}\;\;
    \{(\bm{\alpha}^k, \bm{\beta}^k)\}_{k=1}^K =  \omega(\bm{X}, \bm{A}),
    \end{aligned}
\end{equation}
where $\bm{\alpha}^k,\bm{\beta}^k\in\mathbb{R}^{N\times C}$.
Thus, instead of learning a single parameter shared across all channels, GReLU learns multiple weights to evaluate the importance of different channels. Moreover, adding  $\max(\cdot)$ makes GReLU a piecewise linear function that cuts the input values in different ranges and rescales them differently. This further enables GReLU to consider the effect of different value ranges.

In particular, when $K=1$, GReLU reduces to linear transformations that assign a single weight regardless of the range of input. GReLU  performs similar as  attention mechanism described in global attention~\cite{li2015gated, velivckovic2017graph}. In such a special case, these transformations are also related to Linear Graph Networks SGC~\cite{wu2019simplifying} and S$^2$GC~\cite{zhu2020simple}.

\subsection{Computational Complexity.} 
Below, we analyze the Computational Complexity of GReLU, which is  efficient as we decouple the  node- and channel-wise parameters. Moreover, the computation cost of GReLU is better compared to the multi-head attention mechanism (MA) adopted in GAT. 
\subsubsection{Time Complexity} The complexity of GReLU consists of two parts: (i) the complexity of the hyperfunction $\omega_{(\textbf{X}, \textbf{A})}$ which is used to infer the parameters of GReLU. (ii) the complexity of activation function $\sigma_{\omega_{(\textbf{X}, \textbf{A})}}(\cdot)$ which outputs the activations. 
\renewcommand{\labelenumi}{\roman{enumi}.}
%\vspace{-0.4cm}
\hspace{-1cm}
\begin{enumerate}[leftmargin=0.6cm]
%\item We formulate the FSL problem as to generate dynamic classifiers.
\item  Hyperfunction includes two components: (1) computation of the channel-wise parameters which consists of computing the average of diffusion output and MLP layer (with $\tanh(\cdot)$), and (2) the computation of node-wise parameters which includes the cost of diffusion and one-layer MLP with softmax. In the channel-wise block, GReLU spends  $\mathcal{O}(|\mathcal{E}|)$ in the diffusion layer and $\mathcal{O}(2KC$) in the MLP. In the node-wise block, GReLU spends $\mathcal{O}(|\mathcal{E}|)$ in diffusion and  $\mathcal{O}(|\mathcal{V}|)$ in MLP with softmax.
\item Activation function requires that GReLU is applied per channel of each node, which requires $\mathcal{O}(C|\mathcal{V}|)$ to compute outputs.
\end{enumerate}
Thus, GReLU has the complexity  $\mathcal{O}(|\mathcal{E}|+C|\mathcal{V}|)$. As the computation cost of activation function, $\mathcal{O}(C|\mathcal{V}|$), is the same for all rectifiers, the extra cost is mainly introduced by the computation of hyperfunction. Due to the light design of hyperfunction: 1) the averaging of diffusion output results in a single vector $\bm{e} \in \mathbb{R}^C$; 2) the MLP in node block contains only one output dimension and 3) the choice of $K$ is $2$ (discussed in Section~\ref{sec:keffect}), the computation cost of hyperfunction is negligible compared to cost of GNN.
\subsubsection{Complexity of GReLU \vs Multi-head
Attention Mechanism}
GReLU is faster than the multihead attention mechanism used in GAT. Note that MA needs to compute coefficient for $|\mathcal{E}|$ edges of $M$ heads, resulting in $\mathcal{O}(M|\mathcal{E}|)$, whereas GReLU only computes coefficient for $C$ channel and $|\mathcal{V}|$ nodes, resulting in $\mathcal{O}(|\mathcal{V}|+ C)$ complexity. As  $|\mathcal{V}| >> |\mathcal{E}|$ and $C$ is a relatively small number compared to both $|\mathcal{V}|$ and $|\mathcal{E}|$,  adopting GReLU with a simple backbone (\eg, GCN) is faster than GAT. 

%% file: Content/Experiment.tex
\begin{table*}[t]
  \centering
   \caption{The average node classification accuracy (over 10 runs). The training set and test set are randomly divided. The results in parentheses are the best results achieved among the 10 experiments. We omit the results of Cheb in BlogCatalog and Flickr dataset due to the out-of-memory issue during the training step.}
  \resizebox{0.9\textwidth}{!}
  {
    \begin{tabular}{c|c|ccc|cc|cc}
    \toprule
    \toprule
    % \multicolumn{9}{c} {Node Classification} \\
    % \midrule
    \textbf{Dataset} & \textbf{Model} & \textbf{ReLU}  & \textbf{LReLU} & \textbf{ELU}  &   \textbf{PReLU} & \textbf{Maxout} & \textbf{GReLU} \\
    \midrule
    Cora & Cheb &76.5 ± 2.2(79.4) & 75.5 ± 2.2(80.0) & 74.9 ± 2.9(80.4) & 75.6 ± 3.0(80.6) & 75.4 ± 2.0(77.8) & \textbf{78.3 ± 1.7}(79.4) \\
& GCN &79.2 ± 1.4(81.6) & 79.4 ± 1.6(80.6) & 78.8 ± 1.3(80.2) & 78.9 ± 1.0(80.2) & 79.8 ± 1.5(81.5) & \textbf{81.8 ± 1.8}({83.0})\\
& SAGE &78.6 ± 1.7(81.3) & 77.6 ± 1.8(79.9) & 77.5 ± 1.4(79.9) & 77.6 ± 1.7(78.7) & 77.7 ± 1.7(79.9) & \textbf{79.3 ± 1.5}(80.7) & \\
& GAT &{81.2 ± 1.3}(83.3) & 80.8 ± 2.3(83.1) & 80.5 ± 1.7(83.0) & 80.4 ± 2.1(83.1) & 80.2 ± 1.7(82.5) & \textbf{81.5 ± 2.1}({84.1}) \\
& ARMA &79.0 ± 1.4(80.8) & 79.2 ± 2.1(82.9) & 79.7 ± 0.8(80.6) & 79.3 ± 2.0(81.9) & 79.8 ± 1.3(81.3) & \textbf{80.1 ± 1.5}({82.6}) \\
 &APPNP &{82.0 ± 1.3}(83.1) & 81.0 ± 2.3(83.1) & 81.5 ± 1.7(83.4) & 81.4 ± 2.1(83.4) & 81.0 ± 1.7(82.5) & \textbf{82.5 ± 2.1}({84.7}) \\
    \midrule
    PuMed & Cheb &68.1 ± 4.1(74.9) & 70.5 ± 3.0(76.7) & 68.4 ± 2.8(72.7) & 71.5 ± 3.1(76.3) & 71.8 ± 3.9(75.8) & \textbf{73.4 ± 2.9}(75.6) \\
& GCN &77.6 ± 2.2(81.6) & 76.8 ± 1.6(79.4) & 76.8 ± 2.2(80.5) & 77.3 ± 3.7({82.5}) & 77.3 ± 2.9(80.6) & \textbf{78.9 ± 1.7}(81.3) \\
& SAGE &75.7 ± 3.1(79.8) & 75.3 ± 3.3(79.8) & 74.8 ± 2.7(78.6) & 76.0 ± 2.5(80.5) & 74.5 ± 2.7(77.4) & \textbf{76.2 ± 1.6}(78.4) & \\
& GAT &77.2 ± 3.1(81.3) &{78.7 ± 1.7}({81.2}) & 78.9 ± 2.3(82.1) & 76.2 ± 3.0(80.7) & 77.9 ± 1.7(80.3) &  \textbf{79.1 ± 1.8}(80.6) \\
& ARMA &76.9 ± 2.6(80.7) & 76.5 ± 1.9(80.3) & 77.3 ± 2.5(80.4) & 76.5 ± 2.4(80.7) & 76.6 ± 2.9(81.3) & \textbf{77.4 ± 3.0}(80.2) & \\
 &APPNP &78.2 ± 3.1(82.3) &{78.2 ± 1.7}({81.7}) & 79.0 ± 2.3(82.1) & 79.2 ± 3.0(80.7) & 78.9 ± 1.7(81.3) &  \textbf{79.8 ± 1.8}(81.2) \\
    \midrule
    CiteSeer& Cheb &67.8 ± 1.8(71.0) & 67.1 ± 2.9(71.1) & 67.8 ± 2.4(71.1) & 67.0 ± 2.3(70.5) & 67.4 ± 1.5(70.7) & \textbf{68.1 ± 1.3}(70.4)\\
& GCN &67.7 ± 2.3(72.1) & 68.4 ± 1.8(71.2) & 68.3 ± 1.4(70.1) & 67.3 ± 2.1(70.7) &68.5 ± 2.2({72.5}) & \textbf{68.5 ± 1.9}(71.7)  \\
& SAGE &67.1 ± 2.8(70.1) & 67.3 ± 2.1(70.1) & 67.8 ± 1.7(70.2) & 66.2 ± 2.6(69.6) & 67.5 ± 1.8(71.5) & \textbf{68.0 ± 1.3}(69.7) \\
& GAT &68.6 ± 1.4(70.8) & {69.2 ± 1.9}({71.7}) & 68.4 ± 1.6(71.2) & 68.2 ± 1.6(69.7) & 68.6 ± 1.6(71.3) & \textbf{69.3 ± 1.7}(71.9) \\
& ARMA &67.7 ± 1.3(68.9) & 68.6 ± 2.4(71.5) & 67.9 ± 2.1(71.3) & 66.8 ± 1.5(69.4) & 68.5 ± 1.8(70.9) & \textbf{69.0 ± 2.2}(71.7) \\
 &APPNP &68.7 ± 1.3(70.5) & {69.3 ± 1.6}({71.2}) & 69.4 ± 1.4(71.2) & 69.5 ± 1.6(70.7) & 69.2 ± 1.6(72.0) & \textbf{70.0 ± 1.7}(72.3) \\
    \midrule
    BlogCatalog
          & GCN &72.1 ± 1.9({75.5}) & 72.6 ± 2.1(75.2) & 72.6 ± 1.8(75.3) & 71.4 ± 2.1(74.5) & 72.4 ± 1.4(75.3) & \textbf{73.7 ± 1.2}(74.2) \\
          & SAGE &71.9 ± 1.3({76.0}) & 72.2 ± 1.9(74.9) & 72.1 ± 1.8(74.3) & 72.0 ± 2.1(74.4) & 71.6 ± 1.4(74.3) & \textbf{73.3 ± 1.6}(75.3) \\
          & GAT &41.7 ± 7.3(54.4) & 38.2 ± 3.3(41.8) & 46.6 ± 3.6(51.7) & {67.2 ± 2.6(71.8)} & 54.2 ± 3.9(59.3) & \textbf{67.8 ± 3.9(72.3)}\\
         & ARMA &72.5 ± 3.3(79.1) & 72.5 ± 5.9(78.7) & 77.2 ± 2.2(79.2) & 79.6 ± 3.0(84.5) & 84.4 ± 1.8(86.9) & \textbf{85.7 ± 2.7}{(88.4)} \\
          &APPNP &71.1 ± 1.8({75.3}) & 72.5 ± 2.0(75.2) & 72.4 ± 1.8(75.3) & 71.7 ± 2.1(74.8) & 72.8 ± 1.4(75.3) & \textbf{73.8 ± 1.5}(74.8) \\
    \midrule
    Flickr
         & GCN &50.7 ± 2.3(54.8) & 51.0 ± 2.0(53.8) & 52.8 ± 1.8(56.0) & 53.0 ± 1.6(54.9) & 54.0 ± 1.8(56.8) & \textbf{54.4 ± 1.6}({56.8})\\
          & SAGE &49.7 ± 2.2(53.8) & 50.8 ± 2.1(54.0) & 52.6 ± 1.8(56.7) & 53.2 ± 1.3(56.1) & 54.0 ± 1.7(56.5) & \textbf{55.3 ± 1.4}({57.2})\\
          & GAT &20.2 ± 3.1(25.2) & 20.3 ± 2.5(23.8) & 23.8 ± 2.9(28.1) & {32.8 ± 4.9}({43.2}) & 30.0 ± 2.6(34.6) & \textbf{33.7 ± 3.1}(36.3)\\
          & ARMA &48.4 ± 4.7(56.1) & 52.0 ± 3.9(59.0) & 52.2 ± 4.1(58.1) & 45.9 ± 3.5(53.1) & 52.7 ± 4.0(59.4) & \textbf{56.5 ± 2.4}({59.1})
          \\
           &APPNP &51.6 ± 2.0(54.0) & 51.6 ± 2.1(53.8) & 53.0 ± 1.8(56.0) & 53.2 ± 1.4(54.9) & 53.9 ± 1.9(56.6) & \textbf{54.8 ± 1.9}({56.7})\\
    \bottomrule
    \bottomrule
    \end{tabular}%
    }
 \label{tab:nodecls}%
\end{table*}%
\section{Experiments}
Below, we evaluate the proposed method GReLU on two tasks, node classification in Section~\ref{sec:nodecls} and graph classification in Section~\ref{sec:graphcls}. We show the effectiveness of GReLU by comparing it with different ReLU variants with several GNN backbones. We show the contribution of different modules of GReLU in an ablation study in Section~\ref{sec:ablation}. The effect of choosing the hyper-parameter $K$ is discussed in Section~\ref{sec:keffect}. To verify the GReLU acts adaptively, an analysis of parameters is conducted in Section~\ref{sec:inspect}. We aim to answer the following Research Questions (RQ):
% In the experiment, we evaluate the proposed method in various real-world datasets in the node classification task. We try to answer the following Research Questions (RQ).
% \vspace{-0.3cm}
\begin{itemize}
    \item  \textbf{RQ1: } Does applying GReLU in the updating step help improve the capacity of GNNs for different downstream tasks?
    \item  \textbf{RQ2: } What kind of factors does GReLU benefit from in its design and how do they contribute to performance?
    \item  \textbf{RQ3: } How does the choice of parameter $K$ affect the performance of GReLU?
    \item  \textbf{RQ4: } Does GReLU dynamically  adapt to different nodes?
    
\end{itemize}

\subsection{Node Classification (RQ1)}
\label{sec:nodecls}

\vspace{0.1cm}\noindent\textbf{Datasets.}
% \vspace{0.1cm}\noindent\textbf{Dataset.}  GReLU is evaluated on five real-world datasets which are summarized in Table~\ref{tab:datasets}. The datasets are divided into two major types: (1) Citation Network (\textbf{Cora, PubMed, CiteSeer})~\cite{kipf2016semi} whose nodes are publications and edges are citation links. (2) Social network(\textbf{BlogCatalog} and \textbf{Flickr})~\cite{meng2019co} where edges are extracted from their social relationships and node features are constructed via keywords of user profiles.
GReLU is evaluated on five real-world datasets:
% Their statistics are summarized in Table~\ref{tab:datasets}.
\begin{itemize}
    \item Cora, PubMed, CiteSeer~\cite{kipf2016semi} are well-known citation network datasets, where nodes represent papers and edges represent their citations, and the nodes are labeled based on the paper topics.
    \item Flick~\cite{meng2019co}  is an image and video hosting website, where users interact with each other via photo sharing. It is a social network where nodes represent users and edges represent their relationships, and all nodes are divided into 9 classes according to the interest groups of users.
    \item BlogCatalog\cite{meng2019co}  is a social network for bloggers and their social relationships from the BlogCatalog website. Node attributes are constructed by the keywords of user profiles, and the labels represent the topic categories provided by the authors, and all nodes are divided into 6 classes.
\end{itemize}

\vspace{0.1cm}\noindent\textbf{Baselines.} We compare GReLU with different ReLU variants (as shown in Table~\ref{tab:relus}) and choose several representative GNNs as backbone:
%  We select GCN~\cite{kipf2016semi},  GraphSage~\cite{hamilton2017inductive}, Chebyshev~\cite{defferrard2016convolutional}, GAT~\cite{velivckovic2017graph}, ARMA~\cite{bianchi2019graph}, as our GNN backbones. 
% We summarize ReLU variants in Table~\ref{tab:4} and list GNN as follow:
\begin{itemize}
    \item \textbf{Chebyshev}~\cite{defferrard2016convolutional} is a spectral GNN model utilizing Chebyshev filters.
    \item \textbf{GCN}~\cite{kipf2016semi} is a semi-supervised graph convolutional network model which transforms node representations into the spectral domain using the graph Fourier transform.
    \item \textbf{ARMA}~\cite{bianchi2019graph} is a spectral GNN utilizing ARMA filters.
    \item \textbf{GraphSage}~\cite{hamilton2017inductive} is a spatial GNN model that learns node representations by aggregating information from neighbors.
    \item \textbf{GAT}~\cite{velivckovic2017graph} is a spatial GNN model using the attention mechanism in the aggregate function.
    \item \textbf{APPNP}~\cite{klicpera2018predict} is a spectral GNN that propagates the message via a fully personalized PageRank scheme.
\end{itemize}
\vspace{0.1cm}\noindent\textbf{Experimental Setting.} To evaluate our model and compare it with  existing methods, we use the semi-supervised setting in the node classification task~\cite{kipf2016semi}. We use the well-studied data splitting schema in the previous work \cite{kipf2016semi,velivckovic2017graph} to reduce the risk of overfitting. To get more general results, in Cora, CiteSeer and Pubmed, we do not employ the public train and test split~\cite{kipf2016semi}. Instead, for each experiment, we \textbf{randomly} choose 20 nodes from each class for training and randomly select 1000 nodes for testing. All baselines are initialized with the parameters suggested by their papers, and we also further carefully tune the parameters to get the optimal performance. For our model, we choose $K=2$ for the whole experiment. A detailed analysis of the effect of $K$ is in Section~\ref{sec:keffect}. We use a 2-layer GNN with 16 hidden units for the citation networks (Cora, PubMed, CiteSeer) and 64 units for social networks (Flick, BlogCatalog). In both networks,  dropout rate 0.5 and L2-norm regularization are exploited to prevent overfitting. For each combination of ReLUs and GNNs, we run the experiments 10 times with random partitions and report the average results and the best runs. We use classification accuracy to evaluate the performance of different models.
% \vspace{-0.2cm}

\vspace{0.1cm}\noindent\textbf{Observations.}
The node classification results are reported in Table~\ref{tab:nodecls}. We have the following observations:
\vspace{-0.1cm}
\begin{itemize}
    \item Compared with all baselines, the proposed GReLU generally achieves the best performance on all datasets in different GNNs. Especially, Compared with ReLU, GReLU achieves maximum improvements of 13.2\% on BlogCatalog and 8.1\% on Flickr using ARMA backbone. The results demonstrate the effectiveness of GReLU.
    % \item We also observe that GReLU with GAT backbone is not the best model among other baselines in all datasets. This indicates GReLU is not compatible with GAT. As described in Section~\ref{sec:udstandgrelu}. Both GReLU and GAT can be viewed as ways to weigh different neighborhoods. Coupling modules with the similar functionality is redundant and is easy to cause underfitting. GReLU may over-design for the GAT.
    \item We particularly notice that GReLU with a simple backbone (\ie, GCN) outperforms the GAT with other ReLUs. This indicates that the mask edge attention adopted in GAT is not the optimal solution to weighting the information from the neighbors. As GReLU is both topology- and feature-aware, it can be regarded as a more effective and efficient way to weigh the information over both nodes and channels.
    
    % \item We also notice that
\end{itemize}
%% ************************************
\vspace{-0.3cm}
\subsection{Graph Classification( RQ1)}
\label{sec:graphcls}
\begin{table*}[t]
\caption{Ablation study: different GReLU variants (ABCD) evaluated on the Flickr dataset, the GCN backbone is used.}
    \resizebox{0.8\textwidth}{!}{\begin{tabular}{l|lcll|c}
    \toprule
    \toprule
    \textbf{Model}  & \textbf{Type} & \textbf{K} & \textbf{Hyperfunction} & \textbf{Activation} & Classification Accuracy\\
    \midrule
    GReLU-A & Non-linear & $K=2$ & $\alpha^k_{nc}, \beta^k_{nc} \leftarrow\omega(\bm{X},\bm{A})$ & $\max_{1\leq k \leq K}\{\alpha^k_{nc}x_{nc} + \beta^k_{nc}\}$ & 54.4 ± 1.6\\
    GReLU-B & Non-linear & $K=2$ & $\alpha^k_{nc}, \beta^k_{nc} \leftarrow\omega(\bm{X})$ & $\max_{1\leq k \leq K}\{\alpha^k_{nc}x_{nc} + \beta^k_{nc}\}$ & 53.4 ± 1.8\\
    GReLU-C  & Linear & $K=1$ & $\alpha^k_{nc}, \beta^k_{nc} \leftarrow\omega(\bm{X},\bm{A})$& $\alpha^k_{nc}x_{nc} + \beta^k_{nc}$ & 53.7 ± 1.4\\
    GReLU-D  & Linear & $K=1$ & $\alpha^k_{nc}, \beta^k_{nc} \leftarrow\omega(\bm{X})$& $\alpha^k_{nc}x_{nc} + \beta^k_{nc}$ & 52.2 ± 1.6\\
    \midrule
    ReLU&  Non-linear & / & / & $\max\{\bm{x}_n, 0\}$& 50.7 ± 2.3\\
    SGC &  Linear & / & / &$\bm{x}_n$ & 48.6 ± 1.3\\
    \bottomrule
    \bottomrule
    \end{tabular}%
    }
\label{tab:ab1}
\end{table*}
\begin{table*}[t]
\caption{Ablation study: different GReLU variants (AEFG) are evaluated on the Flickr dataset, the GCN backbone is used.}
    \resizebox{0.8\textwidth}{!}{\begin{tabular}{l|lll|c}
    \toprule
    \toprule
    \textbf{Model}  & \textbf{Parameter} & \textbf{Hyperfunction} & \textbf{Activation } & \textbf{Classification Accuracy}\\
    \midrule
    GReLU-A  & Node/Channel-wise  & $\alpha^k_{nc}, \beta^k_{nc} \leftarrow \omega(\bm{X},\bm{A})$& $\max_{1\leq k \leq K}\{\alpha^k_{nc}x_n + \beta^k_{nc}\}$ & 54.4 ± 1.6\\
    GReLU-E  & Node/Channel-wise &  $\alpha^k_{nc}\leftarrow \omega(\bm{X},\bm{A})$& $\max_{1\leq k \leq K}\{\alpha^k_{nc}x_{nc} \}$ &54.0 ± 1.4\\
    GReLU-F& Channel-wise & $\alpha^k_{c}, \beta^k_{c}\leftarrow\omega(\bm{X},\bm{A})$& $\max_{1\leq k \leq K}\{\alpha^k_{c}x_{nc} + \beta^k_{c}\}$ &53.0 ± 1.9\\
    GReLU-G& Node-wise & $\gamma_{n}\leftarrow\omega(\bm{X},\bm{A})$& $\max_{1\leq k \leq K}\{\gamma^k_{n}x_{n, c}\}$ &53.8 ± 2.1\\
    \bottomrule
    \bottomrule
    \end{tabular}%
    }
  \label{tab:ab2}%
\end{table*}
\begin{table*}[htbp]
  \centering
  \caption{Graph classification results.}
   \resizebox{0.65\textwidth}{!}{
    \begin{tabular}{c|c|ccc|cc|c}
    \toprule
    \toprule
    % \multicolumn{8}{c}{Graph Classification} \\
    % \midrule
    \textbf{Dataset} & \textbf{Model} & \textbf{ReLU}  & \textbf{LReLU} & \textbf{ELU} & \textbf{PReLU} & \textbf{Maxout} & \textbf{GReLU} \\
    \midrule
    {PROTEINS} & GCN   & 76.0 ± 3.2 & 75.1 ± 2.2 & 76.3 ± 3.0 & 76.3 ± 3.2 & 76.5 ± 3.2 & \textbf{76.7 ± 2.8} \\
          & SAGE  & 75.9 ± 3.2 & 75.7 ± 1.2 & 76.5 ± 3.4 & 75.9 ± 3.4 & 76.6 ± 3.2 & \textbf{76.9 ± 1.6} \\
          & GIN   & 76.2 ± 2.8 & 76.1 ± 2.9 & {77.4 ± 2.6} & 76.4 ± 3.8 & 76.8± 2.7 & \textbf{77.8 ± 3.1} \\
    \midrule
    {MUTAG} & GCN   & 85.6 ± 5.8 & 86.2 ± 6.8 & 85.7 ± 5.5 & 86.6 ± 5.3 & 85.9 ± 5.8 & \textbf{87.2 ± 7.0} \\
          & SAGE  & 85.1 ± 7.6 & 84.5 ± 7.7 & 85.1 ± 7.2 & 85.7 ± 7.4 & 86.0 ± 7.6 & \textbf{86.7 ± 6.4} \\
          & GIN   & 89.0 ± 6.0 & 89.2 ± 6.2 & 88.9 ± 5.8 & 89.2 ± 6.1 & 88.7 ± 6.0 & \textbf{89.5 ± 7.4} \\
    \bottomrule
    \bottomrule
    \end{tabular}%
    }
  \label{tab:graphcls}%
\end{table*}%

\vspace{0.1cm}\noindent\textbf{Dataset.} In the graph classification task, our proposed GReLU is evaluated on two real-world datasets MUTAG \cite{xu2018how} and PROTEINS \cite{xu2018how} which are common for graph classification tasks.
% summarized as follows.

% \begin{itemize}
%     \item \textbf{MUTAG}: Citeseer is a research paper citation network, where nodes are publications and edges are citation links. 
%     \item \textbf{Proteins}: Citeseer is a research paper citation network, where nodes are publications and edges are citation links.
% \end{itemize}
 \vspace{0.1cm}\noindent\textbf{Baselines.} Similarly to the node classification task, we compare GReLU with ReLU variants (Table~\ref{tab:relus}) with various GNN-based pooing methods:
 %  To show GReLU can be plugged in different MPNNs. We select GCN~\cite{kipf2016semi}, GIN~\cite{xu2018how}, ASAP~\cite{ranjan2020asap}, SAGPool~\cite{lee2019self}, GlobalAttentionNet ~\cite{li2015gated} and as our backbone MPNNs.
\begin{itemize}
    \item GCN~\cite{kipf2016semi} is a multi-layer GCN model followed by  mean pooling to produce the graph-level representation.
    \item GraphSage~\cite{hamilton2017inductive} is a message-passing model. The max aggregator is adopted to obtain the final graph representation. 
    \item GIN~\cite{xu2018powerful} is a graph neural network with MLP, which is designed for graph classification.
\end{itemize}
 \vspace{0.1cm}\noindent\textbf{Experiment Setting.}
 The graph classification is evaluated on two datasets via 10-fold cross-validation. We reserved one fold as a test set and randomly sampled a validation set from the rest of the nine folds. We use the first eight folds to train the models. Hyper-parameter selection is performed for the number of hidden units $\{16, 32, 64, 128\}$ and the number of layers $\{2, 3, 4, 5\}$ \wrt the validation set. Similarly to the node classification task, we choose $p=0.01$  for LeakyReLU and $K=2$ for our model. We report the accuracy with the optimal parameters setting of each model. The classification accuracy is shown in Table~\ref{tab:graphcls}.
 
 \vspace{0.1cm}\noindent\textbf{Observations.} 
 We observe that  GReLU achieves the best results on both datasets, indicating that the graph-level task can  benefit from the global information extracted by GReLU. Compared to ReLU which is the default setting for most of GNNs, GReLU obtains the maximum improvement of 1.7\% with GCN on PROTEINS. For other backbones,  GReLU also improves the performance.
 Note that the non-linearity choice is orthogonal to the choice of pooling \ie, combining GReLU with high-order pooling \cite{9521687} may yield further improvements on the MUTAG and  PROTEINS datasets.
 %
%% ************************************
\vspace{-0.3cm}
\subsection{Ablation Study (RQ2)}
\begin{figure*}[t]
\minipage[t]{0.28\textwidth}
\includegraphics[width=\columnwidth]{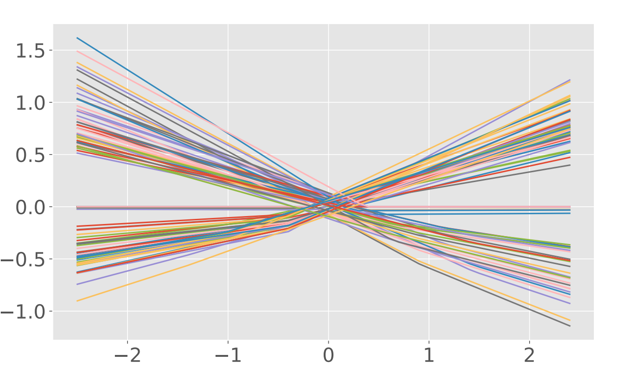}
\caption{The visualization of the GReLU function with $K=2$. Each line in different colors denotes the piecewise function applied on a single channel of each node.}
\label{fig:vis} 
\endminipage
\hspace{5px}
\minipage[t]{0.32\textwidth}
\includegraphics[width=\columnwidth]{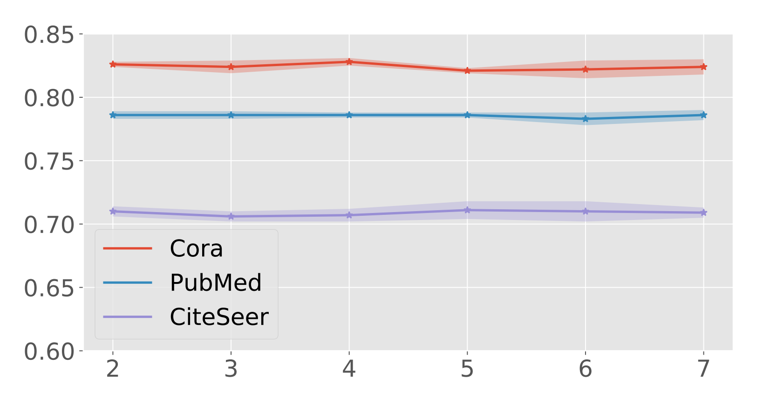}
 \caption{The effect of different $K$ evaluated on Cora, PubMed, and CiteSeer. $x$ axis denotes the number of segments $K$ used in GReLU. $y$ axis denotes the classification accuracy.}
 \label{fig:4} 
 \endminipage\hspace{5px}
 \minipage[t]{0.31\textwidth}
\includegraphics[width=\columnwidth]{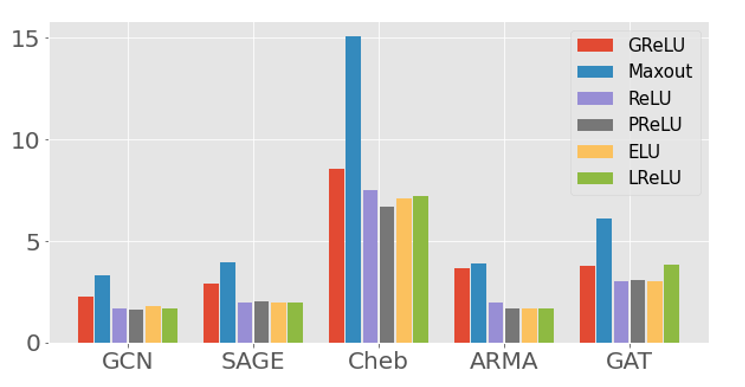}
         \caption{Runtimes of different models with various ReLU variants on CiteSeer. y-axis denotes the runtime of training 200 epochs in seconds.}
          \label{fig:5}
 \endminipage
\end{figure*}
\label{sec:ablation}
Below, we perform the ablation study (Flick dataset) on GReLU by removing different modules  to validate the effectiveness of incorporating the graph topology into a non-linear activation function and the design of hyperfunction of GReLU. To further show the difference between GReLU and previous works, we also compare the variants with SGC. We define four variants as:
\begin{itemize}
    \item GReLU-A: GReLU with $K=2$ which is a non-linear function that rescales the input value in two different ranges.
    \item GReLU-B: No topology information due to omitting the adjacency matrix $\bm{A}$ in the hyperfunction.
    \item GReLU-C: No non-linearity from GReLU-A due to setting $K=1$. In this case, GReLU-C is a linear transformation.
    \item GReLU-D: Both non-linearity and the topology information are removed.
\end{itemize} 
Table~\ref{tab:ab1} shows the difference of GReLU variants and the classification accuracy evaluated on the Flickr dataset. %The observations are as follows. 
The results of GReLU-A and GReLU-D are $1\%$ and $1.5\%$ better than GReLU-B and GReLU-D, indicating the topology information introduced by the adjacency matrix $\bm{A}$ are useful in both linear and non-linear settings. GReLU-A outperforms the GReLU-C, verifying the necessity of adding the non-linearity. Note that  GReLU-B and ReLU use adaptive and non-adaptive rectifiers, respectively.  GReLU-B boosts the classification accuracy from $50.7\%$ to $53.4\%$ which implies the adaptive property plays a vital role in GReLU.

Based on these observations, we conclude that the major gain stems from three parts: introducing the topological structure, adding non-linearity, and making the rectifier be adaptive. Also, as SGC is the GNN without non-linear activation, it is somewhat related to the GReLU with $K=1$. The major difference between them lies in the GReLU adopting the  topology information to reweight the input $x$ using  learnable parameters, whereas SGC uses an identical mapping. We compare SGC with GReLU-C and GReLU-D. Both GReLU variants with or without the adjacency matrix $\bm{A}$ are better than the SGC, which again strengthens our conclusion.

We also study the need for the bias parameter $\beta$ and the use of  node- and channel-wise functions. We further define three  variants:

\vspace{-0.1cm}
%\hfill
\begin{itemize}
    \item GReLU-E: The parameters $\beta$ are dropped.
    \item GReLU-F: The parameters of $\alpha$ and $\beta$ are channel-wise due to removing the node-wise block. All nodes share the same set of parameters.
    \item GReLU-G:  The parameters of $\alpha$ and $\beta$ are node-wise due to removing the channel-wise block. All channels share the same set of parameters.
\end{itemize}

%\hfill
The results are shown in Table~\ref{tab:ab2}. We have observed different degrees of performance degradation by ablating different modules, which confirms the effectiveness of our design.
% \vspace{-0.3cm}
\subsection{The Effect of Parameter $K$ (RQ3)}
\label{sec:keffect} 
% \textcolor{blue}{(allen: detail setting like basic architeture of GNNs)}
Note that $K$, which determines the number of segments in the non-linear function, needs to be predefined for GReLU. In this section, we discuss the effect of $K$ in the GReLU function $\max_{1\leq x\leq K}\{\alpha^k_{nc}x_{nc}+\beta^k_{nc}\}$. We adopt GCN as the backbone model. To show the effect of $K$, we evaluate GReLU on the node classification task \wrt $K$. We choose $K \in \{2, 3, 4, 5, 6, 7\}$. Note that $K=2$ is the default setting for the experiments.  Figure~\ref{fig:4} shows  the fluctuation \wrt different $K$ is small. There is no obvious upward trend and a downward trend for all three datasets. The choice of $K$ is not a significant factor that influences performance. For the sake of saving computations, $K=2$ is recommended.
% \vspace{-0.3cm}
\subsection{Inspecting GReLU (RQ4)}
\label{sec:inspect}
We check if GReLU is adaptive by examining the input and output over different node attributes. We randomly choose 1,000 sets of parameters from GReLU on the node classification task with the  CiteSeer dataset. We visualize the piecewise functions of the GReLU in Figure~\ref{fig:vis} which shows that  they  differ, which indicates GReLU is adaptive to different inputs. Some of the piecewise functions of GReLU are monotonically increasing, indicating both $\alpha_1$ and $\alpha_2$ are positive, which is consistent with other ReLU variants like PReLU and LeakyReLU. We observe that there are functions that are horizontal lines ($y=0$) and lines with small slopes in GReLU. This indicates the parameters ($\alpha_1, \alpha_2, \beta_1, \beta_2$) for a particular channel of a node are zero. As a consequence, there is no activation for this channel $c$ in node $n$.  This shows that GReLU can work as a sparse selector that filters  nodes and channels. Moreover, there are monotonically decreasing piecewise functions indicating $\alpha_1$ and $\alpha_2$ are negative. This is interesting since it will flip the sign and take a controversial effect on its input. Such an effect is similar to the negative part of the concatenated ReLU (CReLU)\cite{shang2016understanding} defined as $\text{CReLU}(x) = [\text{ReLU}(x), \text{ReLU}(-x)].$ CReLU simply makes an identical copy of the linear responses, negates them, concatenates both parts of the activation, and then applies ReLU on both parts. The extra performance is gained by taking ReLU of the negative input $-x$ which is similar to setting  $\alpha < 0$.
% \vspace{-0.3cm}
\subsection{Runtime Comparisons}
%In this section, we compare the running time of GeLU. 
Below, runtimes of training different ReLU variants with 200 epochs on CiteSeer are reported. As shown in Figure~\ref{fig:5}, the runtime of the GReLU is shorter than that of the Maxout and slightly larger compared to other variants. The extra runtime is mainly caused by the hyperfunction that infers the parameters of GReLU, whereas other baselines (ReLU, leakyReLU, ELU, PReLU) do not use any hyperfunction. As discussed in Section~\ref{sec:4}, the hyperfunction of GReLU is simple. GReLU incurs a negligible computation cost.

%% file: Content/Conclusion.tex
\section{Conclusions}
We have proposed a topology adaptive activation function, GReLU, which incorporates the topological information of the graph and incurs a negligible computational cost. The model complexity is controlled by decoupling the parameters from nodes and channels. By making the activation function adaptive to nodes in GNN, the updating functions are more diverse than the typical counterparts in the message passing frameworks. Therefore, GReLU improves the capacity of the GNNs. Although GReLU is designed for the use in the spatial domain with message passing frameworks,  GReLU can be regarded as a plug-and-play component in both spatial and spectral GNNs.

%% file: arxiv.bbl

\begin{thebibliography}{46}

%%% the Web address.

\ifx \showCODEN    \undefined \def \showCODEN     #1{\unskip}     \fi
\ifx \showDOI      \undefined \def \showDOI       #1{#1}\fi
\ifx \showISBNx    \undefined \def \showISBNx     #1{\unskip}     \fi
\ifx \showISBNxiii \undefined \def \showISBNxiii  #1{\unskip}     \fi
\ifx \showISSN     \undefined \def \showISSN      #1{\unskip}     \fi
\ifx \showLCCN     \undefined \def \showLCCN      #1{\unskip}     \fi
\ifx \shownote     \undefined \def \shownote      #1{#1}          \fi
\ifx \showarticletitle \undefined \def \showarticletitle #1{#1}   \fi
\ifx \showURL      \undefined \def \showURL       {\relax}        \fi
% invisible to TeX
\providecommand\bibfield[2]{#2}
\providecommand\bibinfo[2]{#2}
\providecommand\natexlab[1]{#1}
\providecommand\showeprint[2][]{arXiv:#2}

\bibitem[Bianchi et~al\mbox{.}(2019)]%
        {bianchi2019graph}
\bibfield{author}{\bibinfo{person}{Filippo~Maria Bianchi},
  \bibinfo{person}{Daniele Grattarola}, \bibinfo{person}{Lorenzo Livi}, {and}
  \bibinfo{person}{Cesare Alippi}.} \bibinfo{year}{2019}\natexlab{}.
\newblock \showarticletitle{Graph neural networks with convolutional arma
  filters}.
\newblock \bibinfo{journal}{\emph{arXiv preprint arXiv:1901.01343}}
  (\bibinfo{year}{2019}).
\newblock


\bibitem[Chen et~al\mbox{.}(2020a)]%
        {chen2020dynamic}
\bibfield{author}{\bibinfo{person}{Yinpeng Chen}, \bibinfo{person}{Xiyang Dai},
  \bibinfo{person}{Mengchen Liu}, \bibinfo{person}{Dongdong Chen},
  \bibinfo{person}{Lu Yuan}, {and} \bibinfo{person}{Zicheng Liu}.}
  \bibinfo{year}{2020}\natexlab{a}.
\newblock \showarticletitle{Dynamic ReLU}.
\newblock \bibinfo{journal}{\emph{arXiv preprint arXiv:2003.10027}}
  (\bibinfo{year}{2020}).
\newblock


\bibitem[Chen et~al\mbox{.}(2022b)]%
        {chen2021modeling}
\bibfield{author}{\bibinfo{person}{Yankai Chen}, \bibinfo{person}{Menglin
  Yang}, \bibinfo{person}{Yingxue Zhang}, \bibinfo{person}{Mengchen Zhao},
  \bibinfo{person}{Ziqiao Meng}, \bibinfo{person}{Jianye Hao}, {and}
  \bibinfo{person}{Irwin King}.} \bibinfo{year}{2022}\natexlab{b}.
\newblock \showarticletitle{Modeling Scale-free Graphs with Hyperbolic Geometry
  for Knowledge-aware Recommendation}. In \bibinfo{booktitle}{\emph{The
  Fifteenth ACM International Conference on Web Search and Data Mining}}.
\newblock


\bibitem[Chen et~al\mbox{.}(2022a)]%
        {chen2021attentive}
\bibfield{author}{\bibinfo{person}{Yankai Chen}, \bibinfo{person}{Yaming Yang},
  \bibinfo{person}{Yujing Wang}, \bibinfo{person}{Jing Bai},
  \bibinfo{person}{Xiangchen Song}, {and} \bibinfo{person}{Irwin King}.}
  \bibinfo{year}{2022}\natexlab{a}.
\newblock \showarticletitle{Attentive Knowledge-aware Graph Convolutional
  Networks with Collaborative Guidance for Personalized Recommendation}. In
  \bibinfo{booktitle}{\emph{The 38th IEEE International Conference on Data
  Engineering}}.
\newblock


\bibitem[Chen et~al\mbox{.}(2020b)]%
        {IDX}
\bibfield{author}{\bibinfo{person}{Yankai Chen}, \bibinfo{person}{Jie Zhang},
  \bibinfo{person}{Yixiang Fang}, \bibinfo{person}{Xin Cao}, {and}
  \bibinfo{person}{Irwin King}.} \bibinfo{year}{2020}\natexlab{b}.
\newblock \showarticletitle{Efficient community search over large directed
  graphs: An augmented index-based approach}. In
  \bibinfo{booktitle}{\emph{IJCAI}}. \bibinfo{pages}{3544--3550}.
\newblock


\bibitem[Clevert et~al\mbox{.}(2015)]%
        {elus}
\bibfield{author}{\bibinfo{person}{Djork-Arn{\'e} Clevert},
  \bibinfo{person}{Thomas Unterthiner}, {and} \bibinfo{person}{Sepp
  Hochreiter}.} \bibinfo{year}{2015}\natexlab{}.
\newblock \showarticletitle{Fast and accurate deep network learning by
  exponential linear units (elus)}.
\newblock \bibinfo{journal}{\emph{arXiv preprint arXiv:1511.07289}}
  (\bibinfo{year}{2015}).
\newblock


\bibitem[Defferrard et~al\mbox{.}(2016)]%
        {defferrard2016convolutional}
\bibfield{author}{\bibinfo{person}{Micha{\"e}l Defferrard},
  \bibinfo{person}{Xavier Bresson}, {and} \bibinfo{person}{Pierre
  Vandergheynst}.} \bibinfo{year}{2016}\natexlab{}.
\newblock \showarticletitle{Convolutional neural networks on graphs with fast
  localized spectral filtering}.
\newblock \bibinfo{journal}{\emph{Advances in neural information processing
  systems}}  \bibinfo{volume}{29} (\bibinfo{year}{2016}),
  \bibinfo{pages}{3844--3852}.
\newblock


\bibitem[Dugas et~al\mbox{.}(2000)]%
        {softplus}
\bibfield{author}{\bibinfo{person}{Charles Dugas}, \bibinfo{person}{Yoshua
  Bengio}, \bibinfo{person}{Fran{\c{c}}ois B{\'e}lisle},
  \bibinfo{person}{Claude Nadeau}, {and} \bibinfo{person}{Ren{\'e} Garcia}.}
  \bibinfo{year}{2000}\natexlab{}.
\newblock \showarticletitle{Incorporating second-order functional knowledge for
  better option pricing}.
\newblock \bibinfo{journal}{\emph{Advances in neural information processing
  systems}}  \bibinfo{volume}{13} (\bibinfo{year}{2000}),
  \bibinfo{pages}{472--478}.
\newblock


\bibitem[Fu et~al\mbox{.}(2020)]%
        {fu2020magnn}
\bibfield{author}{\bibinfo{person}{Xinyu Fu}, \bibinfo{person}{Jiani Zhang},
  \bibinfo{person}{Ziqiao Meng}, {and} \bibinfo{person}{Irwin King}.}
  \bibinfo{year}{2020}\natexlab{}.
\newblock \showarticletitle{MAGNN: metapath aggregated graph neural network for
  heterogeneous graph embedding}. In \bibinfo{booktitle}{\emph{Proceedings of
  The Web Conference 2020}}. \bibinfo{pages}{2331--2341}.
\newblock


\bibitem[Goodfellow et~al\mbox{.}(2013)]%
        {goodfellow2013maxout}
\bibfield{author}{\bibinfo{person}{Ian Goodfellow}, \bibinfo{person}{David
  Warde-Farley}, \bibinfo{person}{Mehdi Mirza}, \bibinfo{person}{Aaron
  Courville}, {and} \bibinfo{person}{Yoshua Bengio}.}
  \bibinfo{year}{2013}\natexlab{}.
\newblock \showarticletitle{Maxout networks}. In
  \bibinfo{booktitle}{\emph{International conference on machine learning}}.
  PMLR, \bibinfo{pages}{1319--1327}.
\newblock


\bibitem[Hamilton et~al\mbox{.}(2017)]%
        {hamilton2017inductive}
\bibfield{author}{\bibinfo{person}{Will Hamilton}, \bibinfo{person}{Zhitao
  Ying}, {and} \bibinfo{person}{Jure Leskovec}.}
  \bibinfo{year}{2017}\natexlab{}.
\newblock \showarticletitle{Inductive representation learning on large graphs}.
  In \bibinfo{booktitle}{\emph{Advances in neural information processing
  systems}}. \bibinfo{pages}{1024--1034}.
\newblock


\bibitem[He et~al\mbox{.}(2015)]%
        {prelu}
\bibfield{author}{\bibinfo{person}{Kaiming He}, \bibinfo{person}{Xiangyu
  Zhang}, \bibinfo{person}{Shaoqing Ren}, {and} \bibinfo{person}{Jian Sun}.}
  \bibinfo{year}{2015}\natexlab{}.
\newblock \showarticletitle{Delving deep into rectifiers: Surpassing
  human-level performance on imagenet classification}. In
  \bibinfo{booktitle}{\emph{Proceedings of the IEEE international conference on
  computer vision}}. \bibinfo{pages}{1026--1034}.
\newblock


\bibitem[He et~al\mbox{.}(2016)]%
        {he2016deep}
\bibfield{author}{\bibinfo{person}{Kaiming He}, \bibinfo{person}{Xiangyu
  Zhang}, \bibinfo{person}{Shaoqing Ren}, {and} \bibinfo{person}{Jian Sun}.}
  \bibinfo{year}{2016}\natexlab{}.
\newblock \showarticletitle{Deep residual learning for image recognition}. In
  \bibinfo{booktitle}{\emph{Proceedings of the IEEE conference on computer
  vision and pattern recognition}}. \bibinfo{pages}{770--778}.
\newblock


\bibitem[Jarrett et~al\mbox{.}(2009)]%
        {jarrett2009best}
\bibfield{author}{\bibinfo{person}{Kevin Jarrett}, \bibinfo{person}{Koray
  Kavukcuoglu}, \bibinfo{person}{Marc'Aurelio Ranzato}, {and}
  \bibinfo{person}{Yann LeCun}.} \bibinfo{year}{2009}\natexlab{}.
\newblock \showarticletitle{What is the best multi-stage architecture for
  object recognition?}. In \bibinfo{booktitle}{\emph{2009 IEEE 12th
  international conference on computer vision}}. IEEE,
  \bibinfo{pages}{2146--2153}.
\newblock


\bibitem[Kim and Oh(2021)]%
        {DBLP:conf/iclr/supergat}
\bibfield{author}{\bibinfo{person}{Dongkwan Kim} {and}
  \bibinfo{person}{Alice~H. Oh}.} \bibinfo{year}{2021}\natexlab{}.
\newblock \showarticletitle{How to Find Your Friendly Neighborhood: Graph
  Attention Design with Self-Supervision}. In \bibinfo{booktitle}{\emph{9th
  International Conference on Learning Representations, {ICLR} 2021, Virtual
  Event, Austria, May 3-7, 2021}}. \bibinfo{publisher}{OpenReview.net}.
\newblock
\urldef\tempurl%
\url{https://openreview.net/forum?id=Wi5KUNlqWty}
\showURL{%
\tempurl}


\bibitem[Kipf and Welling(2016)]%
        {kipf2016semi}
\bibfield{author}{\bibinfo{person}{Thomas~N Kipf} {and} \bibinfo{person}{Max
  Welling}.} \bibinfo{year}{2016}\natexlab{}.
\newblock \showarticletitle{Semi-supervised classification with graph
  convolutional networks}.
\newblock \bibinfo{journal}{\emph{arXiv preprint arXiv:1609.02907}}
  (\bibinfo{year}{2016}).
\newblock


\bibitem[Klambauer et~al\mbox{.}(2017)]%
        {klambauer2017self}
\bibfield{author}{\bibinfo{person}{G{\"u}nter Klambauer},
  \bibinfo{person}{Thomas Unterthiner}, \bibinfo{person}{Andreas Mayr}, {and}
  \bibinfo{person}{Sepp Hochreiter}.} \bibinfo{year}{2017}\natexlab{}.
\newblock \showarticletitle{Self-normalizing neural networks}.
\newblock \bibinfo{journal}{\emph{Advances in neural information processing
  systems}}  \bibinfo{volume}{30} (\bibinfo{year}{2017}),
  \bibinfo{pages}{971--980}.
\newblock


\bibitem[Klicpera et~al\mbox{.}(2018)]%
        {klicpera2018predict}
\bibfield{author}{\bibinfo{person}{Johannes Klicpera},
  \bibinfo{person}{Aleksandar Bojchevski}, {and} \bibinfo{person}{Stephan
  G{\"u}nnemann}.} \bibinfo{year}{2018}\natexlab{}.
\newblock \showarticletitle{Predict then propagate: Graph neural networks meet
  personalized pagerank}.
\newblock \bibinfo{journal}{\emph{arXiv preprint arXiv:1810.05997}}
  (\bibinfo{year}{2018}).
\newblock


\bibitem[Koniusz and Zhang(2022)]%
        {9521687}
\bibfield{author}{\bibinfo{person}{Piotr Koniusz} {and}
  \bibinfo{person}{Hongguang Zhang}.} \bibinfo{year}{2022}\natexlab{}.
\newblock \showarticletitle{Power Normalizations in Fine-Grained Image,
  Few-Shot Image and Graph Classification}.
\newblock \bibinfo{journal}{\emph{IEEE Transactions on Pattern Analysis and
  Machine Intelligence}} \bibinfo{volume}{44}, \bibinfo{number}{2}
  (\bibinfo{year}{2022}), \bibinfo{pages}{591--609}.
\newblock
\urldef\tempurl%
\url{https://doi.org/10.1109/TPAMI.2021.3107164}
\showDOI{\tempurl}


\bibitem[Koniusz et~al\mbox{.}(2018)]%
        {koniusz2018deeper}
\bibfield{author}{\bibinfo{person}{Piotr Koniusz}, \bibinfo{person}{Hongguang
  Zhang}, {and} \bibinfo{person}{Fatih Porikli}.}
  \bibinfo{year}{2018}\natexlab{}.
\newblock \showarticletitle{A Deeper Look at Power Normalizations}. In
  \bibinfo{booktitle}{\emph{CVPR}}. \bibinfo{pages}{5774--5783}.
\newblock


\bibitem[LeCun et~al\mbox{.}(2015)]%
        {lecun2015deep}
\bibfield{author}{\bibinfo{person}{Yann LeCun}, \bibinfo{person}{Yoshua
  Bengio}, {and} \bibinfo{person}{Geoffrey Hinton}.}
  \bibinfo{year}{2015}\natexlab{}.
\newblock \showarticletitle{Deep learning}.
\newblock \bibinfo{journal}{\emph{nature}} \bibinfo{volume}{521},
  \bibinfo{number}{7553} (\bibinfo{year}{2015}), \bibinfo{pages}{436--444}.
\newblock


\bibitem[Lee et~al\mbox{.}(2020)]%
        {lee2020news}
\bibfield{author}{\bibinfo{person}{Dongho Lee}, \bibinfo{person}{Byungkook Oh},
  \bibinfo{person}{Seungmin Seo}, {and} \bibinfo{person}{Kyong-Ho Lee}.}
  \bibinfo{year}{2020}\natexlab{}.
\newblock \showarticletitle{News Recommendation with Topic-Enriched Knowledge
  Graphs}. In \bibinfo{booktitle}{\emph{Proceedings of the 29th ACM
  International Conference on Information \& Knowledge Management}}.
  \bibinfo{pages}{695--704}.
\newblock


\bibitem[Li et~al\mbox{.}(2015)]%
        {li2015gated}
\bibfield{author}{\bibinfo{person}{Yujia Li}, \bibinfo{person}{Daniel Tarlow},
  \bibinfo{person}{Marc Brockschmidt}, {and} \bibinfo{person}{Richard Zemel}.}
  \bibinfo{year}{2015}\natexlab{}.
\newblock \showarticletitle{Gated graph sequence neural networks}.
\newblock \bibinfo{journal}{\emph{arXiv preprint arXiv:1511.05493}}
  (\bibinfo{year}{2015}).
\newblock


\bibitem[Lim et~al\mbox{.}(2020)]%
        {lim2020stp}
\bibfield{author}{\bibinfo{person}{Nicholas Lim}, \bibinfo{person}{Bryan Hooi},
  \bibinfo{person}{See-Kiong Ng}, \bibinfo{person}{Xueou Wang},
  \bibinfo{person}{Yong~Liang Goh}, \bibinfo{person}{Renrong Weng}, {and}
  \bibinfo{person}{Jagannadan Varadarajan}.} \bibinfo{year}{2020}\natexlab{}.
\newblock \showarticletitle{STP-UDGAT: Spatial-Temporal-Preference User
  Dimensional Graph Attention Network for Next POI Recommendation}. In
  \bibinfo{booktitle}{\emph{Proceedings of the 29th ACM International
  Conference on Information \& Knowledge Management}}.
  \bibinfo{pages}{845--854}.
\newblock


\bibitem[Mairal et~al\mbox{.}(2014)]%
        {ckn}
\bibfield{author}{\bibinfo{person}{J. Mairal}, \bibinfo{person}{P. Koniusz},
  \bibinfo{person}{Z. Harchaoui}, {and} \bibinfo{person}{C. Schmid}.}
  \bibinfo{year}{2014}\natexlab{}.
\newblock \showarticletitle{Convolutional Kernel Networks}.
\newblock \bibinfo{journal}{\emph{NIPS}} (\bibinfo{year}{2014}).
\newblock


\bibitem[Meng et~al\mbox{.}(2019)]%
        {meng2019co}
\bibfield{author}{\bibinfo{person}{Zaiqiao Meng}, \bibinfo{person}{Shangsong
  Liang}, \bibinfo{person}{Hongyan Bao}, {and} \bibinfo{person}{Xiangliang
  Zhang}.} \bibinfo{year}{2019}\natexlab{}.
\newblock \showarticletitle{Co-embedding attributed networks}. In
  \bibinfo{booktitle}{\emph{Proceedings of the Twelfth ACM International
  Conference on Web Search and Data Mining}}. \bibinfo{pages}{393--401}.
\newblock


\bibitem[Misra(2019)]%
        {misra2019mish}
\bibfield{author}{\bibinfo{person}{Diganta Misra}.}
  \bibinfo{year}{2019}\natexlab{}.
\newblock \showarticletitle{Mish: A self regularized non-monotonic neural
  activation function}.
\newblock \bibinfo{journal}{\emph{arXiv preprint arXiv:1908.08681}}
  (\bibinfo{year}{2019}).
\newblock


\bibitem[Nair and Hinton(2010)]%
        {nair2010rectified}
\bibfield{author}{\bibinfo{person}{Vinod Nair} {and}
  \bibinfo{person}{Geoffrey~E Hinton}.} \bibinfo{year}{2010}\natexlab{}.
\newblock \showarticletitle{Rectified linear units improve restricted boltzmann
  machines}. In \bibinfo{booktitle}{\emph{ICML}}.
\newblock


\bibitem[Shang et~al\mbox{.}(2016)]%
        {shang2016understanding}
\bibfield{author}{\bibinfo{person}{Wenling Shang}, \bibinfo{person}{Kihyuk
  Sohn}, \bibinfo{person}{Diogo Almeida}, {and} \bibinfo{person}{Honglak Lee}.}
  \bibinfo{year}{2016}\natexlab{}.
\newblock \showarticletitle{Understanding and improving convolutional neural
  networks via concatenated rectified linear units}. In
  \bibinfo{booktitle}{\emph{international conference on machine learning}}.
  PMLR, \bibinfo{pages}{2217--2225}.
\newblock


\bibitem[Song et~al\mbox{.}(2021a)]%
        {DBLP:conf/cikm/SongMZK21}
\bibfield{author}{\bibinfo{person}{Zixing Song}, \bibinfo{person}{Ziqiao Meng},
  \bibinfo{person}{Yifei Zhang}, {and} \bibinfo{person}{Irwin King}.}
  \bibinfo{year}{2021}\natexlab{a}.
\newblock \showarticletitle{Semi-supervised Multi-label Learning for
  Graph-structured Data}. In \bibinfo{booktitle}{\emph{{CIKM} '21: The 30th
  {ACM} International Conference on Information and Knowledge Management,
  Virtual Event, Queensland, Australia, November 1 - 5, 2021}},
  \bibfield{editor}{\bibinfo{person}{Gianluca Demartini},
  \bibinfo{person}{Guido Zuccon}, \bibinfo{person}{J.~Shane Culpepper},
  \bibinfo{person}{Zi~Huang}, {and} \bibinfo{person}{Hanghang Tong}} (Eds.).
  \bibinfo{publisher}{{ACM}}, \bibinfo{pages}{1723--1733}.
\newblock
\urldef\tempurl%
\url{https://doi.org/10.1145/3459637.3482391}
\showDOI{\tempurl}


\bibitem[Song et~al\mbox{.}(2021b)]%
        {DBLP:journals/corr/zixing}
\bibfield{author}{\bibinfo{person}{Zixing Song}, \bibinfo{person}{Xiangli
  Yang}, \bibinfo{person}{Zenglin Xu}, {and} \bibinfo{person}{Irwin King}.}
  \bibinfo{year}{2021}\natexlab{b}.
\newblock \showarticletitle{Graph-based Semi-supervised Learning: {A}
  Comprehensive Review}.
\newblock \bibinfo{journal}{\emph{CoRR}}  \bibinfo{volume}{abs/2102.13303}
  (\bibinfo{year}{2021}).
\newblock
\showeprint[arXiv]{2102.13303}
\urldef\tempurl%
\url{https://arxiv.org/abs/2102.13303}
\showURL{%
\tempurl}


\bibitem[Srivastava et~al\mbox{.}(2014)]%
        {srivastava2014dropout}
\bibfield{author}{\bibinfo{person}{Nitish Srivastava},
  \bibinfo{person}{Geoffrey Hinton}, \bibinfo{person}{Alex Krizhevsky},
  \bibinfo{person}{Ilya Sutskever}, {and} \bibinfo{person}{Ruslan
  Salakhutdinov}.} \bibinfo{year}{2014}\natexlab{}.
\newblock \showarticletitle{Dropout: a simple way to prevent neural networks
  from overfitting}.
\newblock \bibinfo{journal}{\emph{The journal of machine learning research}}
  \bibinfo{volume}{15}, \bibinfo{number}{1} (\bibinfo{year}{2014}),
  \bibinfo{pages}{1929--1958}.
\newblock


\bibitem[Sun et~al\mbox{.}(2020)]%
        {pmlr-v115-sun20a}
\bibfield{author}{\bibinfo{person}{Ke Sun}, \bibinfo{person}{Piotr Koniusz},
  {and} \bibinfo{person}{Zhen Wang}.} \bibinfo{year}{2020}\natexlab{}.
\newblock \showarticletitle{Fisher-Bures Adversary Graph Convolutional
  Networks}. In \bibinfo{booktitle}{\emph{Proceedings of The 35th Uncertainty
  in Artificial Intelligence Conference}} \emph{(\bibinfo{series}{Proceedings
  of Machine Learning Research}, Vol.~\bibinfo{volume}{115})},
  \bibfield{editor}{\bibinfo{person}{Ryan~P. Adams} {and}
  \bibinfo{person}{Vibhav Gogate}} (Eds.). \bibinfo{publisher}{PMLR},
  \bibinfo{pages}{465--475}.
\newblock
\urldef\tempurl%
\url{http://proceedings.mlr.press/v115/sun20a.html}
\showURL{%
\tempurl}


\bibitem[Veli{\v{c}}kovi{\'c} et~al\mbox{.}(2017)]%
        {velivckovic2017graph}
\bibfield{author}{\bibinfo{person}{Petar Veli{\v{c}}kovi{\'c}},
  \bibinfo{person}{Guillem Cucurull}, \bibinfo{person}{Arantxa Casanova},
  \bibinfo{person}{Adriana Romero}, \bibinfo{person}{Pietro Lio}, {and}
  \bibinfo{person}{Yoshua Bengio}.} \bibinfo{year}{2017}\natexlab{}.
\newblock \showarticletitle{Graph attention networks}.
\newblock \bibinfo{journal}{\emph{arXiv preprint arXiv:1710.10903}}
  (\bibinfo{year}{2017}).
\newblock


\bibitem[Wang et~al\mbox{.}(2020)]%
        {wang2020efficient}
\bibfield{author}{\bibinfo{person}{Yaqing Wang}, \bibinfo{person}{Fenglong Ma},
  {and} \bibinfo{person}{Jing Gao}.} \bibinfo{year}{2020}\natexlab{}.
\newblock \showarticletitle{Efficient Knowledge Graph Validation via
  Cross-Graph Representation Learning}. In
  \bibinfo{booktitle}{\emph{Proceedings of the 29th ACM International
  Conference on Information \& Knowledge Management}}.
  \bibinfo{pages}{1595--1604}.
\newblock


\bibitem[Wu et~al\mbox{.}(2019)]%
        {wu2019simplifying}
\bibfield{author}{\bibinfo{person}{Felix Wu}, \bibinfo{person}{Amauri Souza},
  \bibinfo{person}{Tianyi Zhang}, \bibinfo{person}{Christopher Fifty},
  \bibinfo{person}{Tao Yu}, {and} \bibinfo{person}{Kilian Weinberger}.}
  \bibinfo{year}{2019}\natexlab{}.
\newblock \showarticletitle{Simplifying graph convolutional networks}. In
  \bibinfo{booktitle}{\emph{International conference on machine learning}}.
  PMLR, \bibinfo{pages}{6861--6871}.
\newblock


\bibitem[{Wu} et~al\mbox{.}(2021)]%
        {9046288}
\bibfield{author}{\bibinfo{person}{Z. {Wu}}, \bibinfo{person}{S. {Pan}},
  \bibinfo{person}{F. {Chen}}, \bibinfo{person}{G. {Long}}, \bibinfo{person}{C.
  {Zhang}}, {and} \bibinfo{person}{P.~S. {Yu}}.}
  \bibinfo{year}{2021}\natexlab{}.
\newblock \showarticletitle{A Comprehensive Survey on Graph Neural Networks}.
\newblock \bibinfo{journal}{\emph{IEEE Transactions on Neural Networks and
  Learning Systems}} \bibinfo{volume}{32}, \bibinfo{number}{1}
  (\bibinfo{year}{2021}), \bibinfo{pages}{4--24}.
\newblock
\urldef\tempurl%
\url{https://doi.org/10.1109/TNNLS.2020.2978386}
\showDOI{\tempurl}


\bibitem[Xu et~al\mbox{.}(2015)]%
        {leakyrelu}
\bibfield{author}{\bibinfo{person}{Bing Xu}, \bibinfo{person}{Naiyan Wang},
  \bibinfo{person}{Tianqi Chen}, {and} \bibinfo{person}{Mu Li}.}
  \bibinfo{year}{2015}\natexlab{}.
\newblock \showarticletitle{Empirical evaluation of rectified activations in
  convolutional network}.
\newblock \bibinfo{journal}{\emph{arXiv preprint arXiv:1505.00853}}
  (\bibinfo{year}{2015}).
\newblock


\bibitem[Xu et~al\mbox{.}(2018)]%
        {xu2018powerful}
\bibfield{author}{\bibinfo{person}{Keyulu Xu}, \bibinfo{person}{Weihua Hu},
  \bibinfo{person}{Jure Leskovec}, {and} \bibinfo{person}{Stefanie Jegelka}.}
  \bibinfo{year}{2018}\natexlab{}.
\newblock \showarticletitle{How powerful are graph neural networks?}
\newblock \bibinfo{journal}{\emph{arXiv preprint arXiv:1810.00826}}
  (\bibinfo{year}{2018}).
\newblock


\bibitem[Xu et~al\mbox{.}(2019)]%
        {xu2018how}
\bibfield{author}{\bibinfo{person}{Keyulu Xu}, \bibinfo{person}{Weihua Hu},
  \bibinfo{person}{Jure Leskovec}, {and} \bibinfo{person}{Stefanie Jegelka}.}
  \bibinfo{year}{2019}\natexlab{}.
\newblock \showarticletitle{How Powerful are Graph Neural Networks?}. In
  \bibinfo{booktitle}{\emph{International Conference on Learning
  Representations}}.
\newblock
\urldef\tempurl%
\url{https://openreview.net/forum?id=ryGs6iA5Km}
\showURL{%
\tempurl}


\bibitem[Yang et~al\mbox{.}(2020)]%
        {yang2020featurenorm}
\bibfield{author}{\bibinfo{person}{Menglin Yang}, \bibinfo{person}{Ziqiao
  Meng}, {and} \bibinfo{person}{Irwin King}.} \bibinfo{year}{2020}\natexlab{}.
\newblock \showarticletitle{FeatureNorm: L2 Feature Normalization for Dynamic
  Graph Embedding}. In \bibinfo{booktitle}{\emph{2020 IEEE International
  Conference on Data Mining (ICDM)}}. IEEE, \bibinfo{pages}{731--740}.
\newblock


\bibitem[Yang et~al\mbox{.}(2021)]%
        {yang2021discrete}
\bibfield{author}{\bibinfo{person}{Menglin Yang}, \bibinfo{person}{Min Zhou},
  \bibinfo{person}{Marcus Kalander}, \bibinfo{person}{Zengfeng Huang}, {and}
  \bibinfo{person}{Irwin King}.} \bibinfo{year}{2021}\natexlab{}.
\newblock \showarticletitle{Discrete-time Temporal Network Embedding via
  Implicit Hierarchical Learning in Hyperbolic Space}. In
  \bibinfo{booktitle}{\emph{Proceedings of the 27th ACM SIGKDD Conference on
  Knowledge Discovery \& Data Mining}}. \bibinfo{pages}{1975--1985}.
\newblock


\bibitem[Zhu and Koniusz(2020)]%
        {zhu2020simple}
\bibfield{author}{\bibinfo{person}{Hao Zhu} {and} \bibinfo{person}{Piotr
  Koniusz}.} \bibinfo{year}{2020}\natexlab{}.
\newblock \showarticletitle{Simple spectral graph convolution}. In
  \bibinfo{booktitle}{\emph{International Conference on Learning
  Representations}}.
\newblock


\bibitem[Zhu and Koniusz(2021)]%
        {zhu2021refine}
\bibfield{author}{\bibinfo{person}{Hao Zhu} {and} \bibinfo{person}{Piotr
  Koniusz}.} \bibinfo{year}{2021}\natexlab{}.
\newblock \showarticletitle{Refine: Random range finder for network embedding}.
  In \bibinfo{booktitle}{\emph{Proceedings of the 30th ACM International
  Conference on Information \& Knowledge Management}}.
  \bibinfo{pages}{3682--3686}.
\newblock


\bibitem[Zhu et~al\mbox{.}(2021)]%
        {zhu2021contrastive}
\bibfield{author}{\bibinfo{person}{Hao Zhu}, \bibinfo{person}{Ke Sun}, {and}
  \bibinfo{person}{Peter Koniusz}.} \bibinfo{year}{2021}\natexlab{}.
\newblock \showarticletitle{Contrastive laplacian eigenmaps}.
\newblock \bibinfo{journal}{\emph{Advances in Neural Information Processing
  Systems}}  \bibinfo{volume}{34} (\bibinfo{year}{2021}).
\newblock


\bibitem[Zhu et~al\mbox{.}(2020)]%
        {DBLP:conf/nips/BeyondHomophily}
\bibfield{author}{\bibinfo{person}{Jiong Zhu}, \bibinfo{person}{Yujun Yan},
  \bibinfo{person}{Lingxiao Zhao}, \bibinfo{person}{Mark Heimann},
  \bibinfo{person}{Leman Akoglu}, {and} \bibinfo{person}{Danai Koutra}.}
  \bibinfo{year}{2020}\natexlab{}.
\newblock \showarticletitle{Beyond Homophily in Graph Neural Networks: Current
  Limitations and Effective Designs}. In \bibinfo{booktitle}{\emph{Advances in
  Neural Information Processing Systems 33: Annual Conference on Neural
  Information Processing Systems 2020, NeurIPS 2020, December 6-12, 2020,
  virtual}}, \bibfield{editor}{\bibinfo{person}{Hugo Larochelle},
  \bibinfo{person}{Marc'Aurelio Ranzato}, \bibinfo{person}{Raia Hadsell},
  \bibinfo{person}{Maria{-}Florina Balcan}, {and} \bibinfo{person}{Hsuan{-}Tien
  Lin}} (Eds.).
\newblock
\urldef\tempurl%
\url{https://proceedings.neurips.cc/paper/2020/hash/58ae23d878a47004366189884c2f8440-Abstract.html}
\showURL{%
\tempurl}


\end{thebibliography}
